\documentclass[sigconf]{acmart}

\newcommand{\tabincell}[2]{\begin{tabular}{@{}#1@{}}#2\end{tabular}} 
\usepackage{algorithm}  
\usepackage{algpseudocode}  
\usepackage{amsmath}  
\AtBeginDocument{%
  \providecommand\BibTeX{{%
    \normalfont B\kern-0.5em{\scshape i\kern-0.25em b}\kern-0.8em\TeX}}}

\newcommand{\co}[1]{}

\begin{document}

\title{CGNN: Traffic Classification with Graph Neural Network}

\author{Bo Pang}
\affiliation{%
  \institution{Department of Computer Science and Technology, Harbin Institute of Technology}
  \city{Shenzhen}
  \country{China}
}
\email{20S151084@stu.hit.edu.cn}

\author{Yongquan Fu}
\affiliation{%
  \institution{National University of Defense Technology}
  \city{Changsha}
  \country{China}
}
\email{yongquanf@nudt.edu.cn}

\author{Siyuan Ren}
\affiliation{%
  \institution{Department of Computer Science and Technology, Harbin Institute of Technology}
  \city{Shenzhen}
  \country{China}
}
\email{20S151117@stu.hit.edu.cn}

\author{Ye Wang}
\affiliation{%
  \institution{National University of Defense Technology}
  \city{Changsha}
  \country{China}
}
\email{ye.wang@nudt.edu.cn}

\author{Qing Liao}
\affiliation{%
  \institution{Department of Computer Science and Technology, Harbin Institute of Technology}
  \city{Shenzhen}
  \country{China}
}
\email{liaoqing@hit.edu.cn}

\author{Yan Jia}
\affiliation{%
  \institution{National University of Defense Technology}
  \city{Changsha}
  \country{China}
}
\email{jiayanjy@vip.sina.com}


\begin{abstract}
\co{Graph Neural Network (GNN) has been widely used in node classification, edge classification, etc., and the performance of many GNN algorithms has been proven to be superior to other neural network classification algorithms. Most of today's traffic classification algorithms tend to use CNN, etc., but few people use GNN for traffic classification.} 

Traffic classification associates packet streams with known application labels, which is vital for network security and network management. With the rise of NAT, port dynamics, and encrypted traffic, it is increasingly challenging to obtain unified traffic features for accurate classification. Many state-of-the-art traffic classifiers automatically extract features from the packet stream based on deep learning models such as convolution networks. Unfortunately, the compositional and causal relationships between packets are not well extracted in these deep learning models, which affects both prediction accuracy and generalization on different traffic types.

In this paper, we present a chained graph model on the packet stream to keep  the chained compositional sequence. Next, we propose CGNN, a graph neural network based traffic classification method, which  builds a graph classifier over automatically extracted features over the chained graph. 

Extensive evaluation over real-world traffic data sets,  including normal, encrypted and malicious labels, show that, CGNN improves the prediction accuracy  by 23\% to 29\% for application classification, by 2\% to 37\% for malicious traffic classification, and reaches the same accuracy level for encrypted traffic classification. CGNN is quite robust in terms of the recall and precision metrics. We have extensively evaluated the parameter sensitivity of CGNN, which yields optimized parameters that are quite effective for traffic classification.

\co{In our work, we convert the traffic into a chained graph, the label of each chained graph is the name of the application, each node of the chained graph is the packets, and the feature of the vertex is the packet's content. Then we use GNN to classify the chained graph, and then achieve the purpose of classifying traffic. After testing, we found that the method of using GNN for traffic classification can improve the classification performance.}

\end{abstract}





\maketitle

\section{Introduction}

Traffic classification, which associates traffic clips to known application categories, is one of the building blocks of firewall, intrusion detection, access control, quality of service (QoS) \cite{karakus2017quality} and anomaly detection. Due to the diverse set of applications, the granularity \cite{ZHAO202122} of today's traffic classification is getting finer, and there are more and more traffic types that need to be distinguished by the traffic classifiers. Moreover,  malicious traffic types need to be detected for intrusion detection and prevision in network security. As a result, the classifier should support both normal and malicious traffic.

\co{it is increasingly the demand for high-throughput traffic in today's cyberspace is growing at a rapid rate. Therefore, accurate traffic classification in the current network environment is an challenging and urgent task.}

Traffic classification has received extensive studies. There are typically three classes of classifiers. First, the static feature based methods use static attributes to group traffic clips. For example,  the port-based classification methods classify application types by the ports, e.g.,  HTTP protocol uses port 80, SSL uses port 443. Second, the statistical feature based methods classify pcap object based on traffic metrics. For example, the signature based methods associate each application type with statistical signatures from the traffic samples \cite{2004Accurate}. To adapt to feature distributions, researchers train supervised classifiers based on statistical features from the traffic, such as the flow size, the mean and the standard deviation of the inter-arrival time, the subflow size \cite{CICFlowMeter}. Third, researchers recently directly train the deep neural network models for known application types by learning the features automatically. The neural network model takes multiple layers of transformation over the traffic data, and outputs the prediction by a classification function. 

\co{The training process tries to optimize the loss function of the predicted label to the ground-truth one with known traffic label samples in a supervised manner. The inference is just performing a forward procedure on the neural network for a given traffic object input. }

\textbf{Problem}: Prior traffic classification studies face several severe challenges. First, with the popularity of encrypted traffic and the complexity of advanced networking stacks such as NAT, dynamic ports, it is increasingly challenging to collect useful features for traffic classifier. Second, useful features are increasingly hard to obtain for statistical feature based machine learning-based methods. Manual statistical feature engineering metrics also need more time to calculate the feature summaries for network traffic, which delays the inference time. Moreover, The features need time-consuming handcrafted features that are unsuited for automation, and faces challenges to adapt to drifted  traffic in different network environments such as the WIFI, 5G, industrial internet, or the campus networks, and encrypted traffic also inflate the feature metrics of the original traffic.  As a result, statistical flow metrics alone no longer meet the classification requirements. Third,  DL based classification methods  assume the input to be a fixed Euclidean object, e.g., one-dimensional (1-D) or two-dimensional (2-D) layout structure (like a image), however, the traffic is intrinsically non-euclidean and compositional.

\textbf{Our Work}: In this paper, we address the challenges based on graph neural networks that do not need statistical features, but automatically  capture both structural and semantic relationships based on a chained graph of packets. First, we present a novel graph model for the packet sequence that captures both structural and causal relationships. The graph preserves the linear chain of packet sequence. Second, we present a novel chained graph neural network (CGNN), which automatically extracts and aggregates the traffic attributes from the whole chained graph.

\textbf{High-level Contribution}: The first challenge is to build the chained graph. A straightforward approach is to treat the traffic as a fixed 1-D or 2-D "image". However, due to the interactive nature in the network traffic, the traffic is dynamic and changes among different captured traffic samples, even for the same application type. We capture the interaction process in a chained sequence, and present a chained graph for structure preservation, where the vertex represents the packet, and the edge represents the adjacency relationship between packets.  

The second challenge is to build a traffic-classification oriented machine learning model. Prior studies have adopted convolution networks to improve the traffic classification accuracy. However,  these neural network models do not fit for the compositional chained graph. The graph neural networks, which target for irregular compositional graph models, are promising for the traffic classification problem. Due to the chained structure, we need to preserve the linear chain in the neuron network. Thus, non-linear graph neurons are not fit for the chained graph structure. We adopt the SGC based graph neuron model \cite{8737507} to preserve the linear aggregation relationship in the chained graph.

We evaluate the performance of the traffic classification with real-world data sets, including normal application traffic, malicious traffic, and encrypted traffic. Our evaluation results show that, CGNN improves the prediction accuracy  by 23\% to 29\% for application classification, by 2\% to 37\% for malicious traffic classification, and reaches the same accuracy level for encrypted traffic classification. CGNN is quite robust in terms of the recall and precision metrics. We have extensively evaluated the parameter sensitivity of CGNN, which yields optimized parameters that are quite effective for traffic classification. 

In summary, our main contributions are as follows:
\begin{itemize}
    \item We present a chained graph model to capture the structural and causal relationships in the traffic stream.
    
    \item We present a chained graph neural network model that fit for the structural and causal relationships in the network traffic.
    
    \item We perform extensive evaluation with application traffic, malicious traffic and encrypted traffic, and show that CGNN outperforms state-of-the-art neural network based traffic classifiers.
\end{itemize}

The remainder of this paper is structured as follows. Sec. \ref{ProblemStatementSec} states the traffic classification problem and presents the limitations of existing methods. Sec. \ref{CGSec} presents the chained graph model for traffic classification that captures the structure and causal relationships. Sec. \ref{GNNSec} presents a novel graph neural network on the chained graph for accurate traffic classification. Sec. \ref{TISec} presents the training and inference process for the graph neural network model. Next, Sec. \ref{EvalSec} presents extensive evaluation with real-world data sets. Sec. \ref{RelatedWorkSec} presents prior studies that are most related with our work. Finally, we conclude in 
Sec. \ref{ConclusionSec}.

\section{Problem Statement}
\label{ProblemStatementSec}

\subsection{Traffic Data Model}

\co{The definition of traffic classification is collecting traffic sample, and infer the applications that generate them.} 

We assume that the network traffic data is encoded in the pcap format, which is the standard format for raw packet streams \cite{gharris-opsawg-pcap-02}. The pcap data is a widely used datagram storage format. The pcap fields of each packet consist of two fields: the \textit{header} field that denotes the syntactic metadata of this packet (source/destination IP address, source/destination port, protocol type, etc.); and the \textit{payload} field denotes the semantic information of this packet (application data). Due to the wide adoption of NAT and dynamic ports, we do not consider the IP addresses and the ports.  

\co{, and its file structure is shown in the Figure \ref{pcap} below
\begin{figure}[!t]
  \centering
  \includegraphics[width=\linewidth]{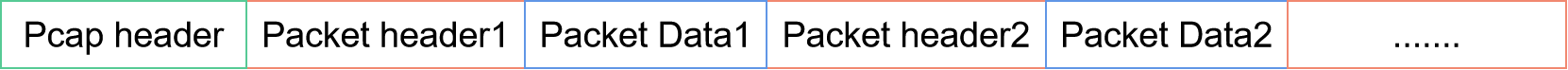}
  \caption{The structure of a pcap file.}
  \label{pcap}
\end{figure}
}

\begin{figure}[!t]
  \centering
  \includegraphics[width=0.4\linewidth]{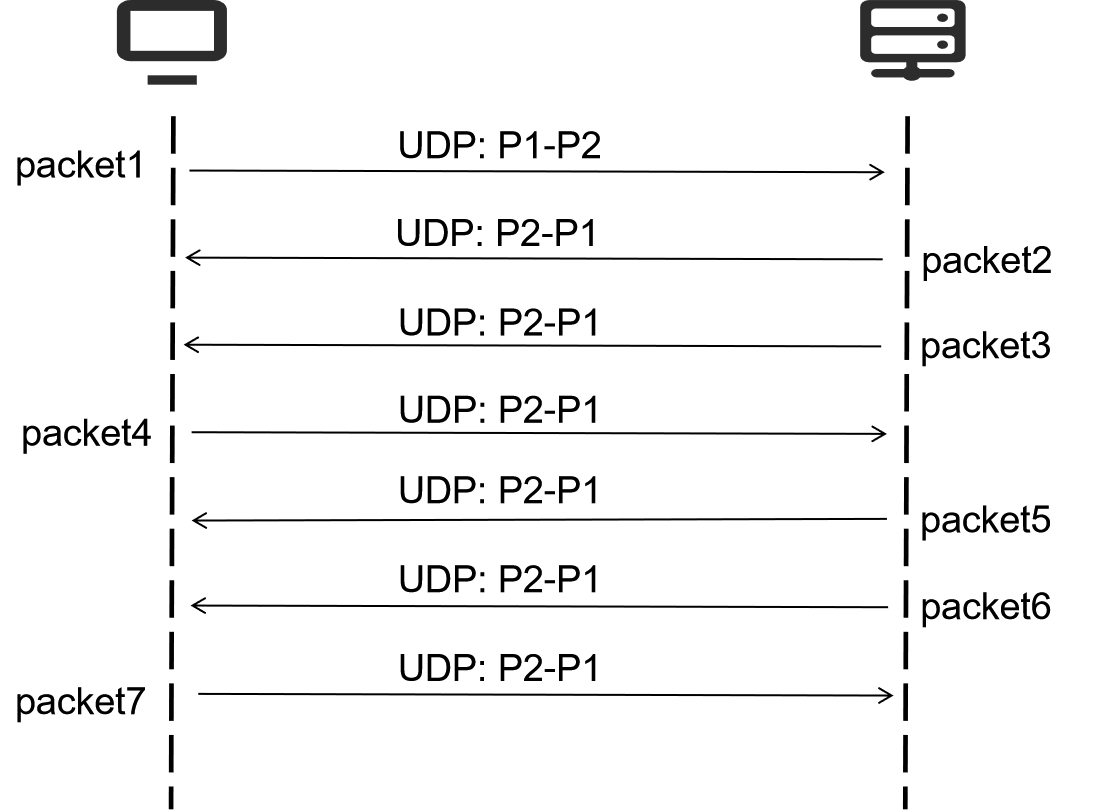}
  \caption{Sequential relationship between packet streams in an application session.}
  \label{img1}
\end{figure}

Specifically, a pcap object  comprises of a sequence of packets that encode the interactions between application endpoints. Figure \ref{img1} shows the time diagram between the packets generated by the Skype software. A typical approach to obtain the pcap object for an application is to collect packets during an active application session period splitted by a duration threshold $\tau$. If there is no  packet transmission within $\tau$ seconds, then we obtain a pcap object associated with an application type. 

\co{The source files we use contains traffic data of a variety of different applications, and each type of application data contains multiple pcap files. Our classification problem is to predict which application the pcap file belongs to based on the content and timing relationship of these pcap files.}

\subsection{Traffic Classification Approaches and Challenges}


The port-based classification methods may be the most straightforward approach, which classify application types by the ports, e.g.,  HTTP protocol uses port 80, SSL uses port 443. When the port changes due to NAT or the dynamic port, these methods do not suffice. 

Different from the static attribute based approaches, the statistical feature based approaches define statistic feature metrics and calculate them from the traffic, and learn the classification model in a supervised approach. For example, researchers which associate each application type with statistical signatures from the traffic samples \cite{2004Accurate}. The statistical features need time-consuming handcrafted features that are unsuited for automation, and faces challenges to adapt to evolved traffic, since   the application signature may dynamically change.


Recently, researchers have  proposed to adopt the deep learning (DL) techniques for traffic classification. The neural network model takes multiple layers of transformation over the traffic data, and outputs the prediction by a classification function. The training process tries to optimize the loss function of the predicted label to the ground-truth one with known traffic label samples in a supervised manner. The inference is just performing a forward procedure on the neural network for a given traffic object input.  Although recent DL based classification methods have promising performance, they still face both structural and causal mismatches: (i) \textit{Structure mismatch}: DL models typically  assume the input to be a fixed one-dimensional (1-D) or two-dimensional (2-D) layout structure (like a image), however, the traffic is a chained compositional structure, and a traffic sample may just represent a clip of the whole interaction process. Consequently, the fixed-structured DL models do not exploit such chained and compositional process. (ii) \textit{Causal mismatch}: Besides the structure mismatch, traditional DL methods learn either a neural network (NN) model for a single packet, or an NN model over the ``image'' of the pcap object, which do not match the causal relationships between the packet sequence.

\subsection{Overview of Our Approach}

Having presented the weakness of existing traffic classification methods, we next propose CGNN, a graph neural network based traffic classification method, that preserves the chained compositional sequence with a semantic-rich graph neural network model. 

First, we present a new graph representation for pcap files. The challenge here is that the pcap file has a variable size, which requires a unified representation model. To that end, we propose a chained graph, where the vertex represents a packet, and the edge between two vertices represents the happen-before relationship of two packets in the pcap file. As a result, we preserve both the time sequence in traffic, and the causal relationship for the network activities.  

Second, we next present a chained graph neural network CGNN for identifying the application type of a pcap file. A CGNN has two aggregation layers over the chain graph neighborhood, one pooling layer over the whole graph,  and one output layer to classify the application type. CGNN does not need complex feature-engineering techniques with a simple structure for performent implementation. The graph neural network structure includes 4 layers, the first and second layers use the SGC \cite{wu2019simplifying} models, the third layer uses AVGPooling \cite{2013Network} layer, and the output layer is a fully connected layer.
Our results on normal, malicious and encrypted data shows that CGNN is not only more accurate than state-of-the-art neural network based classification methods and shows a higher degree of expressiveness and the transfer capability.

\co{An undirected chained graph is used to model the transmission sequence process of network flow, and the message is taken as the vertex and the transmission sequence relationship of the message is taken as the edge to design the extraction vector that adapts to the characteristics of the network flow. This method can fully mine and maintain the characteristic information of the transmission sequence of the network stream.

Secondly, the graph neural network is used. The network architecture constructs a network application identification model, and classifies the chained diagrams describing the sequence structure of traffic and traffic information.}

\section{Chained Graph}
\label{CGSec}

We next present the chained graph that captures the causal relationship for the raw traffic data. The chained graph provides a unified representation model for traffic classification based on the graph neural network. 

\subsection{Motivation}

A unified model capturing the traffic sequential information is currently missing for the traffic data.
Among the existing deep learning-based traffic classification methods, the CNN-based method cannot use the sequential relationship as the feature information. The RNN-based methods can take the sequential relationship into account. However, their characteristics are relatively simple, and the payload of the flow information can not combined with the sequential information. 

Our key insight is that, network traffic can be transformed into a graph structure to capture both structure and semantics of the traffic data. Traffic have been considered as graphs in prior studies \cite{DBLP:conf/imc/IliofotouPFMSV07}. However, prior studies focus on graph mining for a large scale of network addresses, while our goal is to perform machine learning on the packet sequence that typically consists of just two network addresses (source and destination). As a result, we need a unified graph model to capture the structure and the feature semantics of the packet sequence.

\co{, for example, the flow data is transformed into an image \ref{9213630}, or different endpoints in the network are used as vertices, and their mutual exchanges are used as edges to generate a network flow graph \ref{busch2021nf}. }

\subsection{Pcap Preprocessing}
\label{Preprocessing}

Pcap files typically mix multiple application sessions, and some packets do not refer to application sessions, which are noisy for classification. As a result, we need to preprocess the pcap file to keep essential information for traffic classification. 


First, we  divide each pcap file to sessions identified by each 5-tuple. Each session refers to a unique tuple of source IP address, source port number, destination IP address, destination port number, protocol type, and corresponds to a new pcap file. Each session is stored into a new pcap file.


Second, we clean the packets in the pcap file from the first step before converting these pcap files into a chained graph: (i) The Ethernet header contains information about the physical link, which is meaningless for traffic classification. Therefore,  we remove the Ethernet header, in order to avoid interference from irrelevant information and improve the effectiveness of feature extraction. (ii) The transport header contains information about the transport protocol for classification. The length of the TCP header is 20 bytes and the length of the UDP header is 8 bytes. Thus we pad zeros at the end of the UDP header to ensure 20 bytes.  (iii) The IP header contains IP addresses and associated IP protocol. As the IP addresses are not meaningful, we remove the IP address from the IP header. (iv) Some packets do not carry payloads, e.g., those with  SYN, ACK, FIN indicator, and some DNS packets. These packets are not application oriented, thus  they can be safely discarded.  After preprocessing operations on the  pcap file, each packet is converted into a fixed-length vector. we next generate the chained graph based on processed pcap file.

\subsection{Chained Graph Model}

A chained graph refers to preprocessed Pcap session that associates with an application category. 
The vertex corresponds to a packet in the pcap file, while the edge refers to the adjacency relationship between a pair of packets in the pcap file. Each vertex is associated with a feature vector that refers to the  1500-byte vector for this packet. Each edge is undirected by default to increase the degree of message exchanges for the neural network training. Each vertex has a feature vector of length 1500, and the feature vector of $n$ vertices generates an $n\ast1500$ feature matrix.

After obtaining the vertices, we then extract the set of edges between the vertices according to the adjacency storage sequence between the packets. If the two original messages are adjacent in the traffic file, an undirected edge is established between the corresponding two vertices. Here we use undirected edges because undirected edges can better capture the relative sequence relationship between packets than directed edges. The use of undirected edge connection between vertices also allows each vertex to know the information of its predecessor and ancestor packets.

\begin{figure}[!t]
  \centering
  \includegraphics[width=\linewidth]{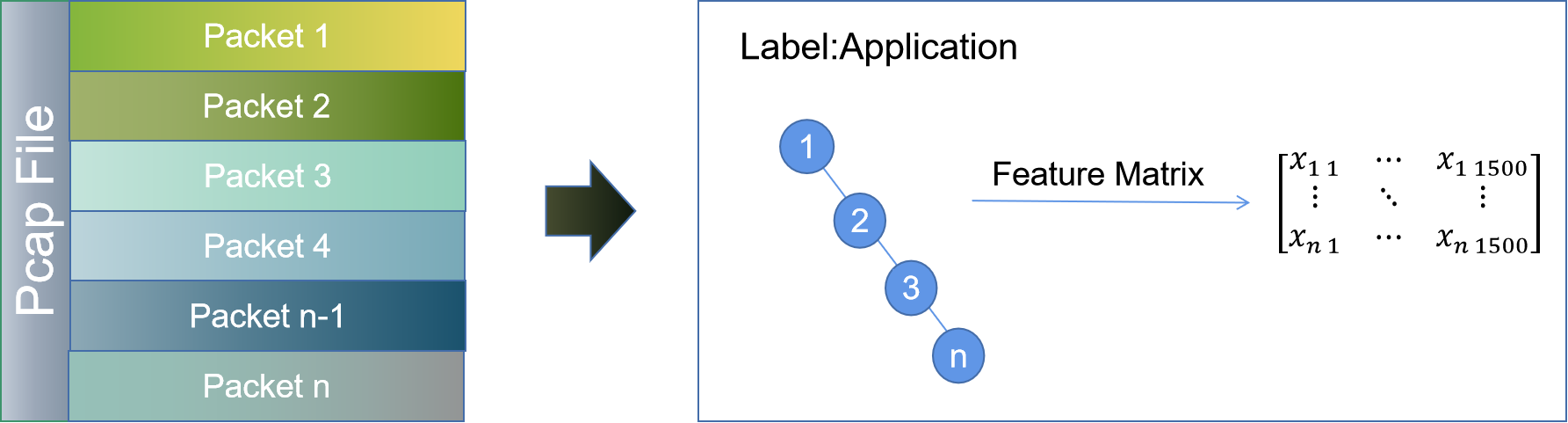}
  \caption{Chained graph model.}
  \label{pcap2graph}
\end{figure}

Finally, we next assign the application type to this graph structure as its label. The original pcap file is converted into a chained graph structure.  Figure \ref{pcap2graph} shows an example of the chained graph model. Generally, a pcap file with $n$ packets corresponds to a chained graph with $n$ vertices and $n-1$ edges. Each packet refers to a vertex, and each edge concatenates a pair of adjacent vertices. A vertex associates with the packet information as its feature vector. The label of this chained graph is its application.

\co{The vertex set of the chained graph contains each packet in the traffic file, that is, we let the packet be the vertex of the chained graph. According to the foregoing,

 of the traffic file. 

In summary, combined with the structure information of the chained graph, we convert a pcap file into a chained graph, as shown in 

, a feature matrix of the chained graph, and a label of the chained graph
}


\co{
the detailed diagram of the chained graph structure and its vertex feature vector is shown in Figure \ref{img5}. 
\begin{figure}[!t]
  \centering
  \includegraphics[width=\linewidth]{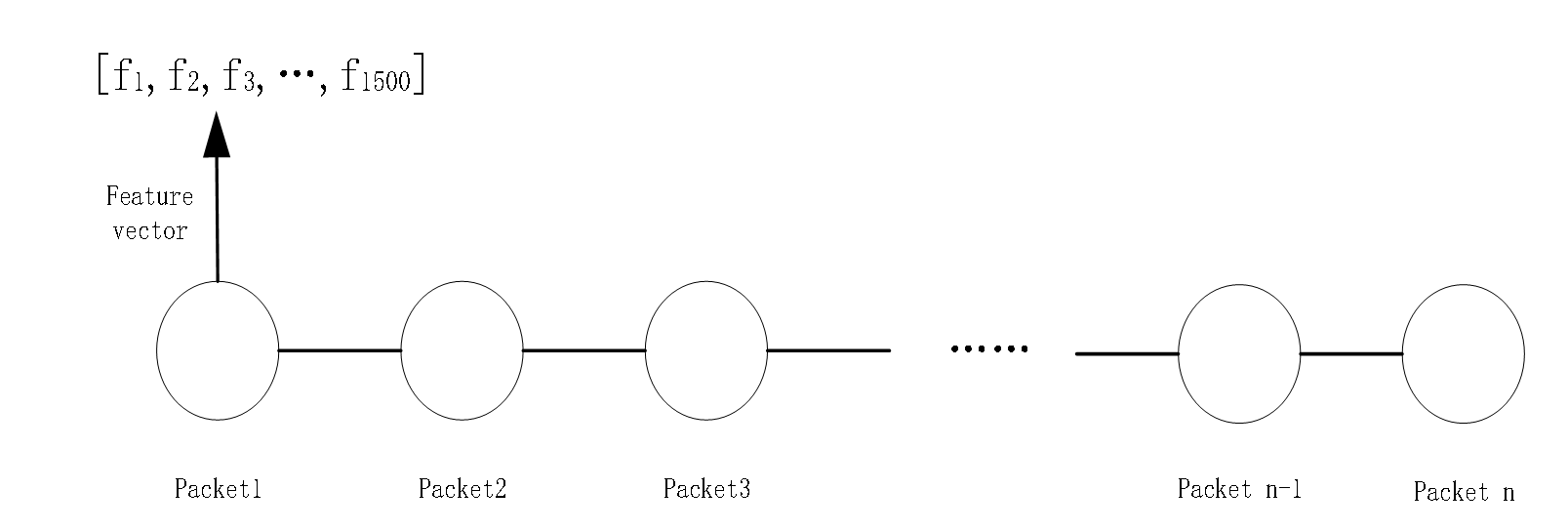}
  \caption{The chained graph of a pcap file.}
  \label{img5}
\end{figure}
}

\subsection{Graph Properties}

The chained graph has several distinct attributes that should be considered for traffic classification: (i) Regular-degree: The degree of each vertex is quite regular due to the chained structure: The first and the last vertices have just one neighbor, while the other ones have two neighbors. (ii)  Long-diameter: The diameter of the chained graph amounts to the number of edges, due to the chained structure. (iii) Causality: Each edge captures the causal relationship between packets, since it associates with the ordering sequence in the interaction between application endpoints. (iv) Feature-preserving: Each vertex is associated with the raw packet's bit level full information.

\section{Graph Neural Network on chained graph}
\label{GNNSec}



In the previous chapter, we introduced how to convert the raw pacp traffic files into chained graphs one by one. Now these chained graphs will replace the raw traffic data as the input of the CGNN model. In this chapter, we will describe the chained graph neural network (CGNN) for accurate traffic classification.

\subsection{Design Choices for Graph Neural Networks}

The properties of the chain graph pose interesting challenges to the choice of the graph neural networks. First, which kind of graph neurons works on the chain graph? Second, how many layers are necessary to obtain competitive performance? Third, how to aggregate the vertex vectors to obtain the representation for the whole chain graph?

First, we need to extract vertex-level features based on the graph neuron. The challenge is to preserve the linear structure in the chained graph. Therefore,   non-linear graph neurons are not the best choices for our graph structure. Instead,  the SGC based graph neuron model \cite{wu2019simplifying}, which performs linear feature aggregation between different layers, preserves the linear semantics in the chained graph. 

After extracting the vertex-level features, the next challenge is to obtain the graph-level feature that summarize the global features of the chained graph. Further, different feature dimensions may contribute a varying degree of semantics.  We capture the global feature by averaging the feature vectors of different vertices, which adapts to different graph structures. Next, we account for different feature dimensions based on a fully connected layer.

Finally, we need to classify the graph-level feature to a traffic category. To that end, we calculate the classification result based on  the softmax classifier.

\begin{figure}[!t]
  \centering
  \includegraphics[width=\linewidth]{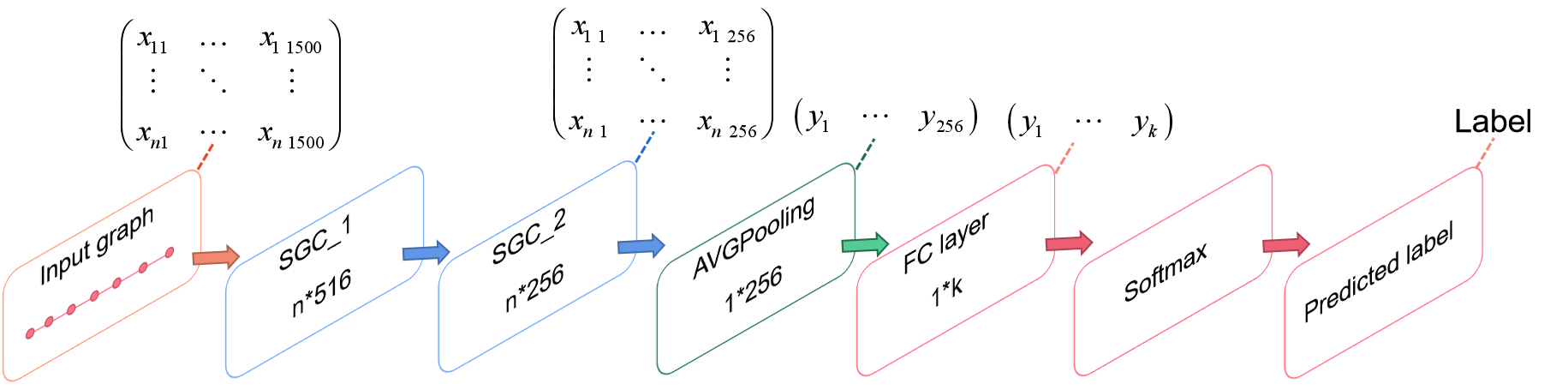}
  \caption{CGNN structure. In the figure, $k$ represents the number of types of traffic in a data set.}
  \label{img6}
\end{figure}

The CGNN structure is shown in Figure \ref{img6}. The input of the CGNN model is the chained graph generated by a pcap file. A CGNN model builds a sparse neural network structure on the chained graph, which aggregates feature vectors of the vertices along the chained structure, and outputs the classification probability for this chained graph.

We take an empirical approach to determine the best parameters for CGNN similar to most graph neural network studies. Our parameter sensitivity analysis in in Subsection \ref{SenseSec} are as follows: (i) Different graph neurons have varying prediction accuracy, due to the diverse adaption to the chain structure. (ii) The SGC graph neuron obtains the highest accuracy among evaluated models. This is mainly because it captures the chain structure in a unified framework. (iii) Two layers of graph neurons are sufficient to achieve near optimal performance, which is consistent to prior graph neural network models. (iv) An average based pooling operator achieves the highest prediction accuracy among evaluated graph pooling operators. (v) The graph neural network has a degree of transfer capability from normal traffic to malicious or encrypted traffic.



\subsection{Graph Model}

\begin{table}[!t]
  \caption{CGNN's key notations.}
  \label{Variable}
  \begin{tabular}{cl}
    \toprule
    Variable &description\\
    \midrule
    $p$ & \tabincell{c}{Dimension of the preprocessed packet}\\
    $m$ & \tabincell{c}{Number of classificiation categories}\\
    $d_1$ & \tabincell{c}{Dimension of the first layer's vector}\\
    $d_2$ & \tabincell{c}{Dimension of the second layer's vector}\\
    $[{{\theta}}^{(1)},{{\theta}}^{(2)}, {{\theta}}^{(3)}]$ & \tabincell{c}{CGNN's weight parameters}\\ 
    \bottomrule
  \end{tabular}
\end{table}

We next present the model specification of each layer in CGNN. Table \ref{Variable} explains the variables in GCNN. Let $n$ represent the number of vertices, $A=[a_{ij}]$ represents the adjacency matrix of the chained graph, and $a_{ij}$ represents whether the vertex $i$ is adjacent to the vertex $j$ (1 represents adjacent, Zeros  represent non-adjacent relationship). Let $D=diag(\sum_j{a_{1j}},..., \sum_j{a_{nj}})$ represent the degree of each vertex as a diagonal matrix (the $i$-th diagonal element is $\sum_j{a_{ij}}$, and the non-diagonal element is 0). $I$ is the identity matrix. Let $\tilde{A}=A+I$, $\tilde{D}$ represents the vertex degree diagonal matrix of the adjacency matrix $\tilde{A}$, where the $i$-th diagonal element of $\tilde{D}$ is $\sum_j{{\tilde{a}}_{ij}}$. Let $S=\tilde{D}^{-\frac{1}{2}}\tilde{A}^{-\frac{1}{2}}\tilde{D}^{-\frac{1}{2}}$, $\theta^{(k)}$ represents a parameter Matrix, where $k$ represents the index of the parameter matrix. Let $X=[x_1,...,x_n ]^T$ denote the matrix of the feature vectors of all vertices, which is of size $n \times p$, where $p$ denotes the default size of the packet. The default size $p$ of each feature vector $x_i$ is 1500 bytes.

A CGNN model consists of four layers.  The first layer is built based on the SGC model. The   input is the feature matrix $X$.  A   single-layer SGC model structure of the first layer is expressed as:  $\overline{X}^{(1)}=SX\theta^{(1)}$.  The model outputs a feature matrix ${\overline{X}}^{(1)}$, where the size $d_1$ of the feature vector of each vertex is 516 bytes by default.

The second layer is similar to the first layer, which also uses a single-layer SGC model over the input  $\overline{X}^{(1)}$. Similar to the first layer, the single-layer SGC model structure of the second layer is expressed as:$\overline{X}^{(2)}=S{\overline{X}^{(1)}}\theta^{(1)}$. The output is a feature matrix $\overline{X}^{(2)}$, where the size $d_2$ of the feature vector of each vertex is 256 bytes by default. 

The third layer aggregates the vertex-level feature matrix to obtain the vector for the whole graph. We choose the AVGPooling operator to obtain the graph-level representation. The input is the feature matrix $\overline{X}^{(2)}$, and the output vector $y$ is the feature vector of the entire chained graph, where $y^{(i)}={\frac{1}{N_i}}\sum_{k=1}^{N_i}{\overline{X}^{(2)}}_k^{(i)}$, where $N_i$ represents the number of vertices in the chained graph and ${\overline{X}^{(2)}}_k^{(i)}$ represents the $i$-th feature of the $k$-th vertex in the chained graph.

The output layer uses a fully connected layer model to obtain the classification result. The input $y$ passes through a linear transformation function $y^{(2)}=W^{T}y+b$, where $W$ represents the weight parameter, and the $b$ represents bias term. Let ${{\theta}}^{(3)} = (W^{T}, b)$ denote the weight parameters that need to be optimized. The   length of the output vector  $y{(2)}$ amounts to  the number of classification categories. Next, we calculate the classification result based on the softmax function over the output vector $y^{(2)}$: $\hat{Y}=softmax(y^{(2)}) $, where $softmax(x_i)=\frac{e^{x_i}}{\sum_{c=1}^Ce^{X_c}}$, where $C$ represents the number of classified categories.

We dimension the CGNN model based on experiments. Specifically, we set the input dimension of the first layer SGC to 1500, which is the length of the feature vector of a packet, and the output dimension is 516, based on the experiments. We set the input dimension of the second-level SGC model to 516, which is the length of the output feature vector of the first-level SGC, and the output dimension is 256. We set the input dimension of the AVGPooling model is $n\times256$, where $n$ represents the number of vertices, and the output is $1\times256$.

\subsection{Example}

\begin{figure}[!t]
  \centering
  \includegraphics[width=\linewidth]{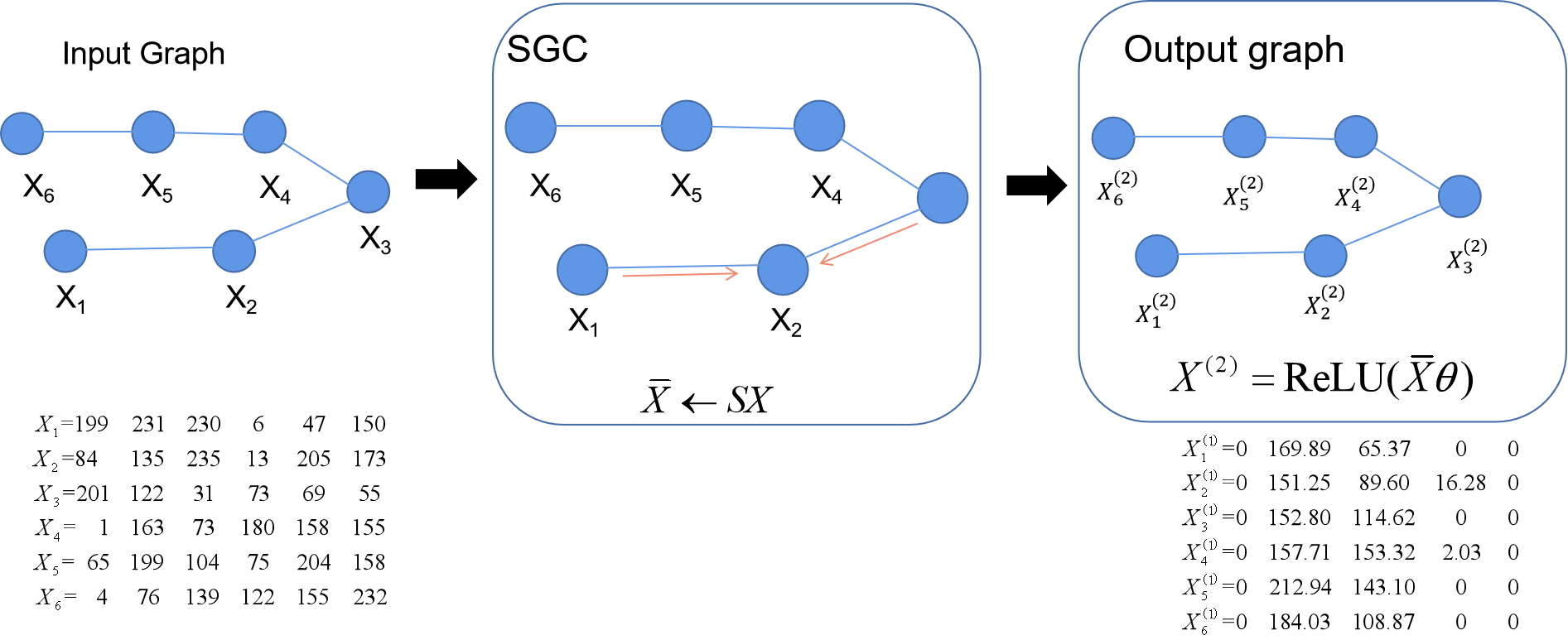}
  \caption{Diagram of the first level SGC.}
  \label{FirstSGC}
\end{figure}
\begin{figure}[!t]
  \centering
  \includegraphics[width=\linewidth]{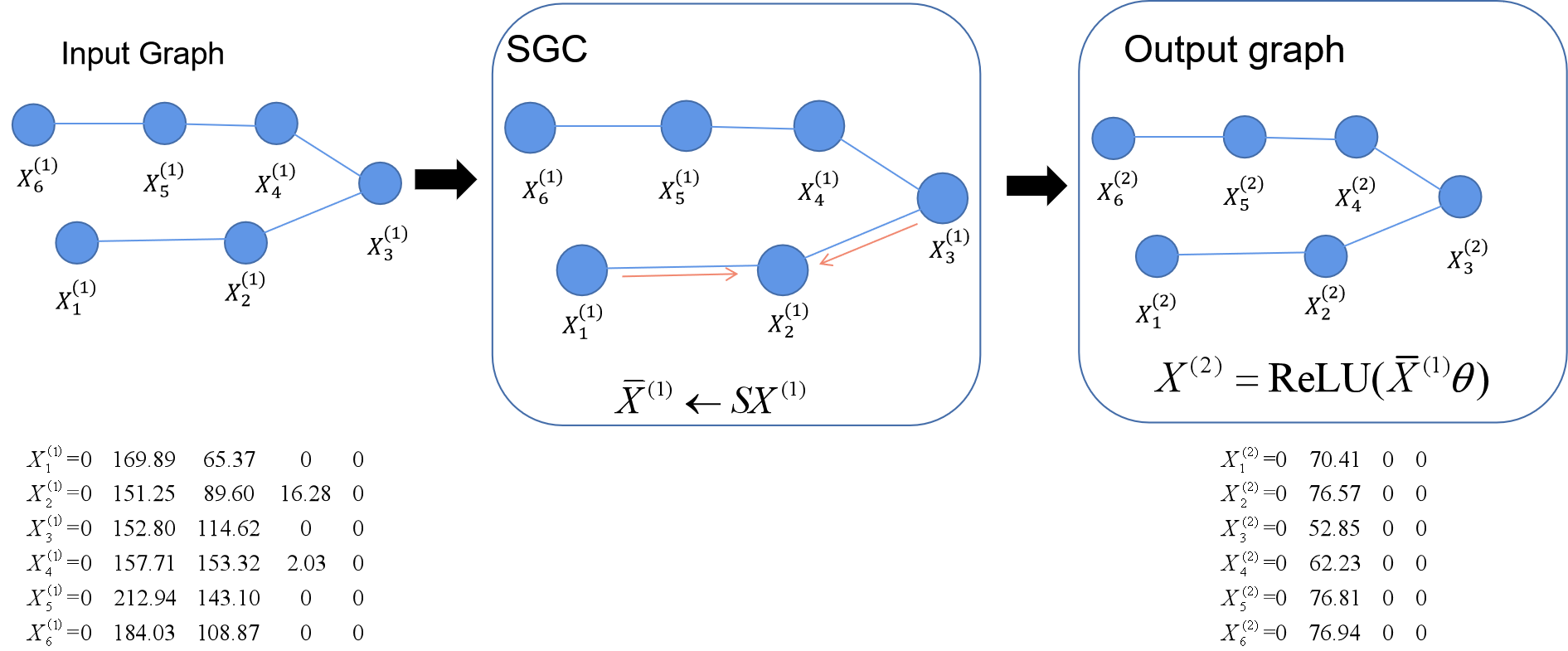}
  \caption{Diagram of the second level SGC.}
  \label{SecondSGC}
\end{figure}
\begin{figure}[!t]
  \centering
  \includegraphics[width=\linewidth]{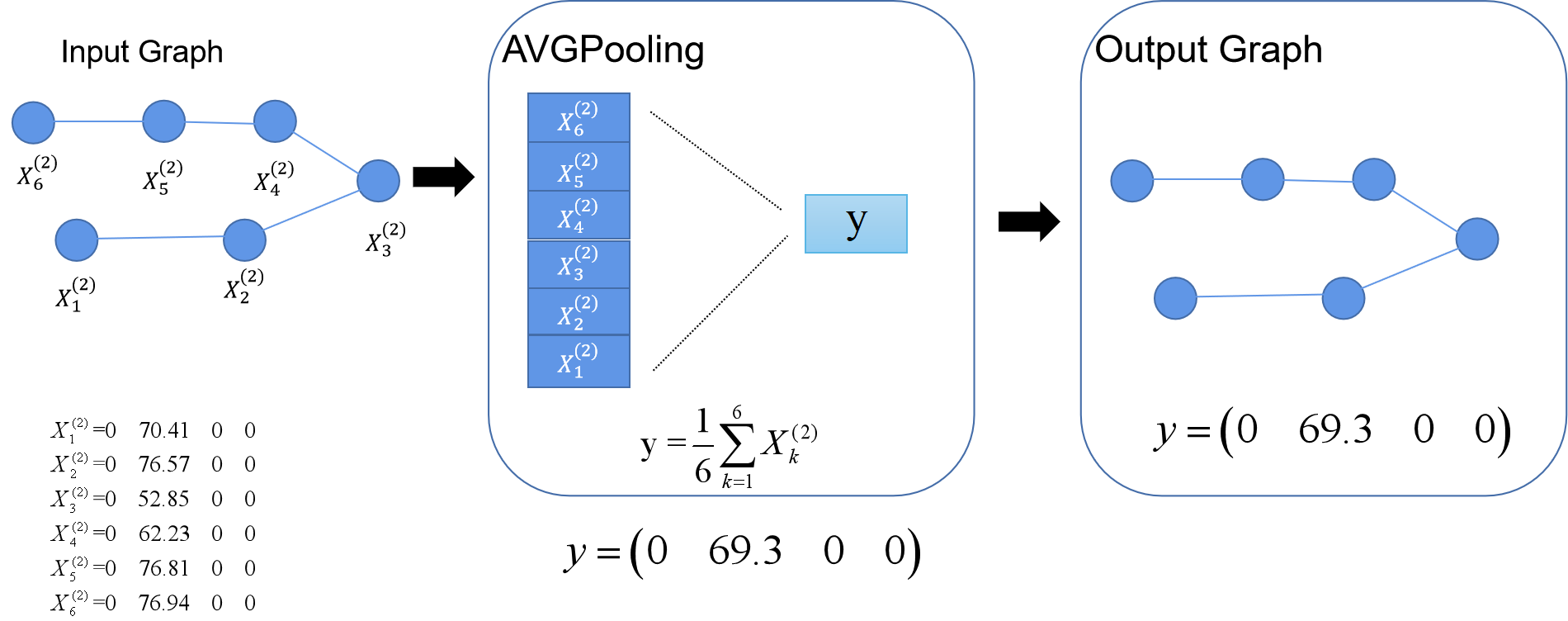}
  \caption{Diagram of the AVGPooling layer.}
  \label{AVGPooling}
\end{figure}
\begin{figure}[!t]
  \centering
  \includegraphics[width=\linewidth]{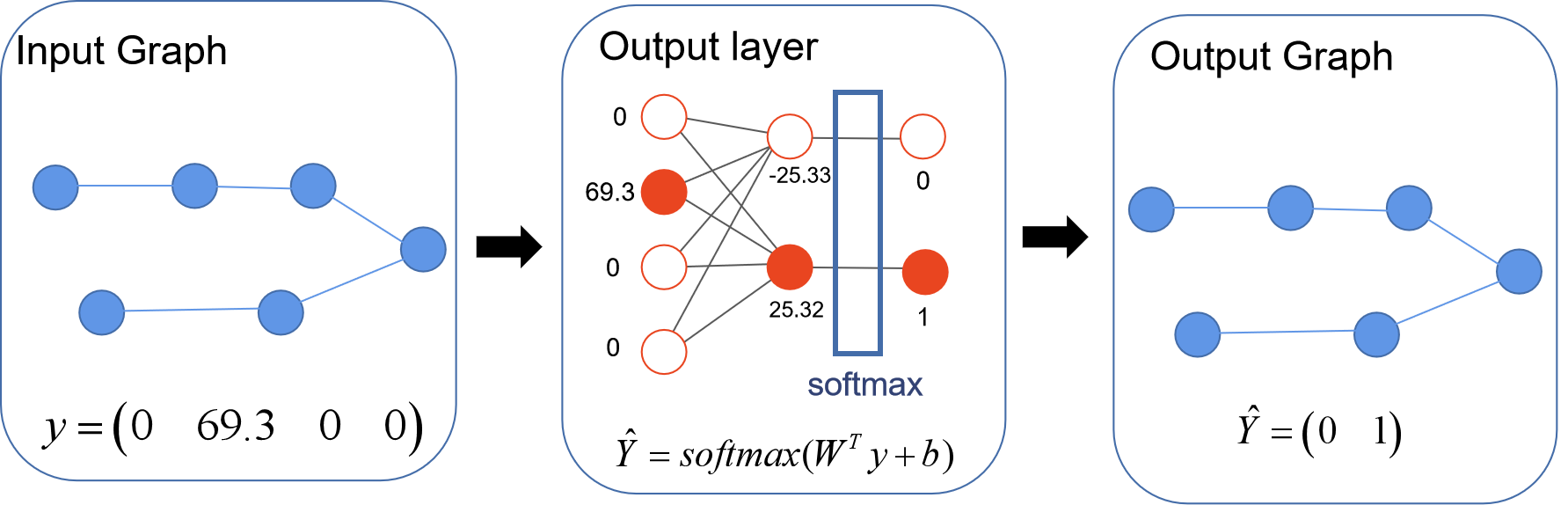}
  \caption{Diagram of the Output layer.}
  \label{OutputLayer}
\end{figure}

We next introduce the CGNN model with an example. The example is for illustration purpose only. Here we assume that a traffic file contains 6 packets, thereby generating a chained graph with 6 vertices. We use this chained graph to simulate the working process of the CGNN model. In this example, the number of input features of the first layer of SGC is 6, and the output is 5. The number of input features of the second layer of SGC is 5, and the number of output features is 4. The number of input features of the fully connected layer is 4, and the number of output features is 2. And in this example, the number of categories is 2, that is, the chained graph can only be classified into the first or second category. The CGNN's layer by layer operations are illustrated from Figure \ref{FirstSGC} to Figure \ref{OutputLayer}. 

The first and second layers aggregate the features on the chained graph. First, Figure \ref{FirstSGC} shows the processing of the chained graph by the first-level SGC. The input of the first layer of SGC is a $6\times6$ feature matrix. After SGC calculation, a $6\times5$ feature matrix is obtained. The feature vector of each row is the feature vector of a vertex in the graph. Second,  the second layer of SGC obtains a $6\times4$ feature matrix, as shown in Figure \ref{SecondSGC}, which is similar to the first layer.

The third layer obtains the graph-level representation by pooling the features of different vertices. Figure \ref{AVGPooling} illustrates the effect of the AVGPooling layer on features. The obtained $6\times4$ feature matrix is pooled, and the $6\times1$ feature vector is obtained as the feature vector of the entire graph. 

The fourth layer takes the graph feature vector as the input, and uses a fully connected layer to obtain a $2\times1$ feature vector. Finally, the output of the fourth layer is fed to the softmax function to obtain the classification result. Figure \ref{OutputLayer} shows the running process of the entire output layer. It can be seen that the length of the $\hat{Y}=[0,1]$ vector after softmax normalization is 2 and the value of the second item is 1, indicating that the probability that this chained graph belongs to the second class is 100\%. With this probability, we can predict that the category of the input chained graph is 2, and so on to get the prediction results of all chained graphs.






\subsection{Spotlights of CGNN}

\textbf{Structural extraction}: CGNN adapts to the regular-degree property through an SGC graph neuron operator that preserves the simplicity of the chained graph. Further, it adapts to the long-diameter chained structure via two layers of automatic aggregation between neighbors of neighbors. 

\textbf{Semantic extraction}: CGNN exploits the causality of the chained sequence to obtain the feature vector via multiple layers of aggregation between chained neighbors.  CGNN preserves the  full semantic information of each packet through the feature vectors of each packet. It extracts the graph-level feature vector through the pooling operator of all feature vectors.

\textbf{Efficiency}: The ultra-sparse chained graph structure enables fast aggregation operations. Further,  CGNN consists of just two layers of simple SGC operators and one layer of pooling operator to obtain the feature vector of the whole chained graph. These operators do not involve time-consuming operations.

\section{CGNN Learning}
\label{TISec}

Having shown the CGNN framework, we next present the training and inference processes for the CGNN model.

\co{We can directly obtain the classification result for the input chained graph, however, the prediction accuracy will be unsatisfying, since the model has not been optimized towards the chained graph samples. Fourth, w}

\textbf{Training}: We generate a chained graph from the traffic datasets. Second, we extract the feature vector sequence of the diagram $[x_1,...,x_n]^T$ from the traffic datasets. Third, we initialize the CGNN model. We train the CGNN model with chained graph samples, and obtain the CGNN parameters. 

The training process for the CGNN seeks to minimize the loss function of the traffic classifier between the estimated label and the ground-truth label. The training function is trained with samples of chained graphs. We use the logistic regression loss function to construct the minimized multi-category loss function. We minimize the loss function with the Adam optimizer. The sample batch training size is set to 32 by default, the maximum number of training rounds is 500, and the network application recognition model represented is output after training. The details can be found in Appendix Algorithm \ref{alg:Framwork}.

\co{Input the traffic file to be tested into the network application recognition model, and output the corresponding network application category. }

\textbf{Inference}: After the training process, we can directly predict the application type with the input chained graph based on the optimized CGNN model $M$. First, the test input is sent in to get $\overline{X}_t^{(1)}$ through the first layer of SGC neural network. Then the output of the first layer serves as the input for the second layer's SGC operator, and the output is $\overline{X}_t^{(2)}$. Next, the output of the second layer serves as the input for the  AVGPooling layer, the output graph feature set $Y_t^{(1)}=[y_1^{(1)},...,y_m^{(1)}]$ is obtained. Then, the output of the third layer passes through the final layer, and the prediction result $\hat{Y}_t=[\hat{y}_1,...,\hat{y}_m$ of the chained graph in the test set is obtained. The details of the inference procedure can be found in Appendix Algorithm \ref{alg:InferenceFramwork}.
 

{\bfseries Time complexity:} The model contains two layers of SGC, a pooling layer and a fully connected layer. Let $m$ denote the number of classification categories. Then the time complexity of the inference for a chained graph is $O((p^{2}+d_1^{2}+p+d_1)\times n+(2m+1)\times d_2+(2m-p-d_1))$ = $O(np^2 + nd_1^2+(n-1)p +(n-1)d_1 +(2m+1) d_2)$.

{\bfseries Space complexity:} The space complexity of the model is related to the number of neurons set, and the space complexity obtained in the CGNN model is $O((p^{2}+d_1^{2}+p+d_1)\times n+(2m+1)\times d_2+2m)$. 


\section{Evaluation}
\label{EvalSec}

Having presented the GNN based classification method, we next compare its performance with state-of-the-art deep learning methods on real-world data sets. 

We implement the training and inference processes for the CGNN model based on the Deep Graph Library (DGL) \cite{wang2019dgl} with Pytorch3.7 \cite{NEURIPS2019_9015} as the backend. 

\subsection{Data Sets}

We would like to test whether CGNN generalizes to different types of traffic. Thus we select three kinds of traffic datasets including application dataset, malicious dataset, and the encrypted dataset. The characteristic data of the three types of data sets used are shown in the table\ref{tab:commands}. In these three types of datasets, the application dataset and the malicious dataset are extracted by ourselves from a real-world network testbed. We manually label each kind of traffic category. The encrypted dataset uses the public data set ISCX \cite{2016Characterization}.  More detailed analysis of the data sets can be found in the appendix. 

\co{The model we propose can be applied to normal traffic data (such as QQ music, Skype applications), and can also be used to distinguish malicious traffic (such as Trojan, Exploit) and encrypted traffic. Therefore, the data sets we have used are mainly divided into common application data sets, malicious traffic data sets and encrypted traffic data sets. The raw data storage form of these three datasets are all pcap files.}

\begin{table}[!t]
  \caption{Dataset characteristics}
  \label{tab:commands}
  \begin{tabular}{ccccc}
    \toprule
    Dataset &Pkt length    & Size   &Pcap files   &Label categories\\
    \midrule
    Application  & 857 & 870M & 1280 & 41\\
    Malicious  & 777 & 76M & 217 & 5\\
    Encrypted  & 127 & 1.2G & 11 & 11\\
    \bottomrule
  \end{tabular}
\end{table}

\begin{figure}[!t]
  \centering
  \includegraphics[width=0.9\linewidth]{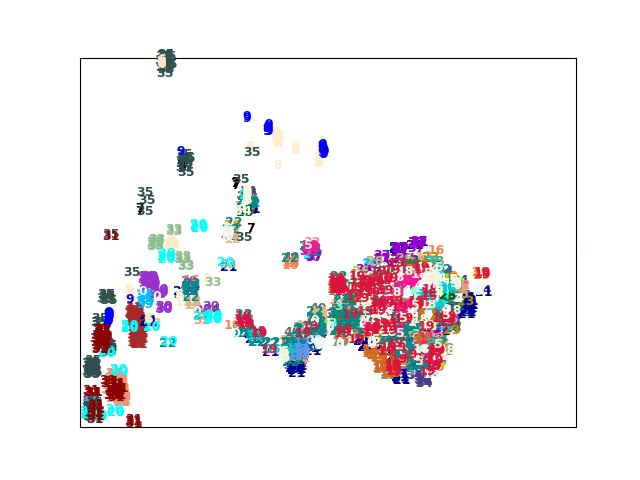}
  \caption{Use t-SNE to visualize the characteristics of the application data set.}
  \label{app_t_sne}
\end{figure}

{\bfseries Application traffic dataset.} This data set contains traffic data generated by network applications. These traffic data are generated by 41 different network applications, so this data set contains 41 different network application traffic data. As mentioned earlier, we need to preprocess the pcap files first. Then we divide the preprocessed pcap files into training set, validation set and test set according to the ratio of 8:1:1. In order to express the discrimination between the 41 application traffic characteristics contained in the application data set, we used t-Distributed Stochastic Neighbor Embedding (t-SNE) dimensionality reduction technology \cite{A2015Visualizing}. Figure  \ref{app_t_sne} shows the degree of difference between the characteristics of application traffic data. We input each preprocessed pcap message, and each message will generate a feature vector after preprocessing. We use t-SNE dimensionality reduction technology to convert these feature vectors formed by the packets into 2-dimensional vectors. In the figure, a kind of  traffic data will be identified by a color. Therefore, it can be seen that there is a certain degree of distinction between the 41 kinds of application traffic data, but there are also certain similarities.

\begin{figure}[!t]
  \centering
  \includegraphics[width=0.9\linewidth]{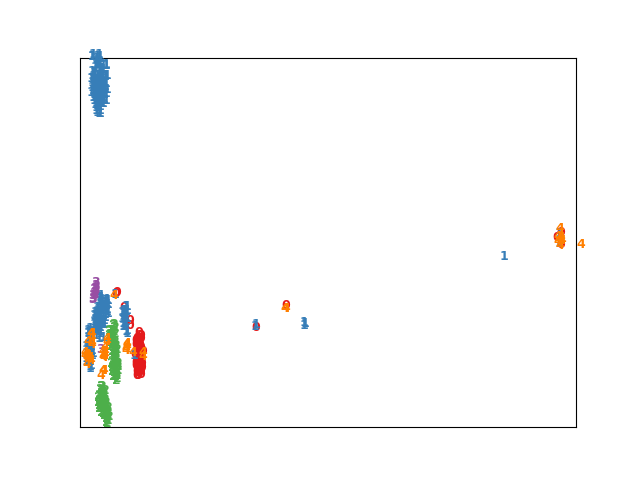}
  \caption{Use t-SNE to visualize the characteristics of malicious data traffic by dimensional reduction.}
  \label{t_sne}
\end{figure}

{\bfseries Malicious traffic dataset.} This data set contains malicious traffic data in traffic. This data set contains 5 different types of malicious traffic data, we also need to preprocess them, and divide the preprocessed pcap files into training set, validation set and test set according to the ratio of 8:1:1. As shown in the Figure \ref{t_sne}, there is a certain degree of distinguishability among the five data sets of malicious traffic. In the figure, each color represents an malicious traffic.

\begin{figure}[!t]
  \centering
  \includegraphics[width=0.9\linewidth]{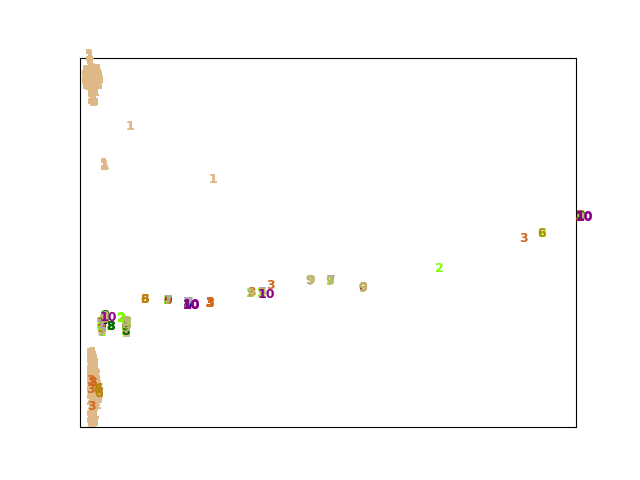}
  \caption{Use t-SNE to visualize the characteristics of encrypted traffic by dimensional reduction.}
  \label{entropy_t_sne}
\end{figure}

{\bfseries Encrypted traffic dataset.} This data set uses the "ISCX VPN-nonVPN" traffic data set. This data set contains 15 types of encrypted traffic, including the encrypted traffic of various types of software such as dating software and video software. After preprocessing the pcap files of encrypted traffic, we will divide the encrypted traffic data set into training set, validation set and test set according to the ratio of 8:1:1. Figure \ref{entropy_t_sne} shows the similarities and distinctions among the 11 types of encrypted traffic in the encrypted traffic data set. In the figure, each color represents an encrypted traffic.

\subsection{Experimental Setup}
\label{paramSet}

The method proposed in this paper is mainly to analyze the data content of traffic data and then classify it. The comparison method used is also a classification method that uses a deep learning model to analyze the payload. So there are four baseline methods used in this paper: 1D-CNN \cite{2017End}, 2D-CNN \cite{2017Malware}, Deep Packet \cite{2020Deep}, and FS-Net\cite{8737507}. Both the 1D-CNN and 2D-CNN models trim the raw pcap files in the data set and convert them into PNG images, then use Convolutional Neural Network (CNN) to classify the PNG images to complete the classification of those traffic data. The Deep Packet method uses a deep learning model to detect and classify the payload of the traffic. The FS-Net method uses the length of the packets as the feature, and uses multi-layer bi-GRU to encode and decode the feature to obtain the traffic classification result.

{\bfseries Experimental parameters:} In this paper, we choose to keep the first 1500 bytes of the packet. If the packet has less than 1500 bytes, we will fill it with 0 to 1500 bytes. In order to train the neural network more efficiently, we pack multiple samples into mini-batch graphs. In the model, we package every 32 chained graphs in the training set into a small batch and send them to the classification model for training. To prevent data overfitting, we used a modified early stopping technique \cite{2012Early}. In order to train the graph neural network model, we used the cross-entropy loss function and also used Adam as the optimizer. The upper limit of training rounds is 400. In order to accelerate the classification of graphics, the batch image classification algorithm in dgl is used in the training process, where the batch size is 32. Finally, the two graph neural network layers in the proposed model use Rectified Linear Unit (ReLU) as the activation function. 

{\bfseries Performance metrics:} In order to compare the performance of the classifier, we use three indicators: Recall(Rc), precision(Pr), Accuracy(Ac), their mathematical description is as follows:
\begin{equation}\label{eq2}
    Rc=\frac{TP}{TP+FN}, Pr=\frac{TP}{TP+FP}, Ac=\frac{TP+FN}{TP+FN+TN+FP}
\end{equation}
where TP, FP, FN and TN stands for true positive, false positive, false negative and true negative, respectively. 

{\bfseries Comparison Methods:} We compare our method with several state-of-the-art neural network model based application methods:

\begin{enumerate}
\item Deep Packet,
Deep Packet \cite{2020Deep} classifies the content of encrypted traffic, and inputs the payload part of the traffic data as a feature vector to the classifier. The classifier is composed of a two-layer convolutional neural network (CNN), a MaxPooling layer and a three-layer fully connected layer. The result of the classification is the predicted application label.
\item 1D-CNN, 
1D-CNN \cite{2017End} is a new end-to-end encrypted traffic classification method based on convolutional neural network. This method uses raw traffic data as the feature. Unlike Deep Packet, it uses the first 785 bytes of the message as the feature, while Deep Packet uses the first 1500 bytes as the feature. The 1D-CNN model converts the trimmed raw pcap files into PNG images. It is then classified using a one-dimensional convolutional neural network (CNN). 
\item 2D-CNN,
2D-CNN \cite{2017Malware} is a traffic classification model that classifies malicious traffic. It also uses CNN as a representation learning technology to send the original data of the traffic data as features to the classifier. The data processing method of 2D-CNN is the same as that of 1D-CNN. The original pcap files are trimmed and converted into PNG images. The difference is that 2D-CNN subsequently uses a two-dimensional convolutional neural network to classify images.

\item FS-Net,
FS-Net \cite{8737507} is an encrypted traffic classification model. Unlike the previous three classification models, it does not use the content of the packet as the features, but uses the length of the packets as the features. The embedded features are successively sent to a multi-layer bi-GRU encoder and a multi-layer bi-GRU decoder to obtain the processed features, which are combined for classification.
\end{enumerate}

\subsection{Comparison Results}

We use 5 methods (CGNN, Deep Packet, 1D-CNN, 2D-CNN,FS-Net) to conduct comparative experiments on application traffic data sets, malicious traffic data sets and encrypted traffic data sets. For the CGNN model, we use the early stopping technique, and the maximum number of epochs is set to 400. In the Deep Packet classification model, two layers of CNN, one layer of MaxPooling and three layers of fully connected layers are used, the maximum number of epochs is 300, and the early stopping technology is used. The $25\times1$ convolution kernel is used in 1D-CNN, and the $5\times5$ convolution kernel is used in 2D-CNN. FS-Net uses a multi-layer dual GRU encoder to learn the representation of a stream sequence, and uses a multi-layer dual GRU decoder to reconstruct the original sequence. The number of epochs set in FS-Net is 40000. And CGNN's classification for encrypted traffic has reached a level of accuracy that exceeds most methods.

{\bfseries (i) Experimental results of application traffic: }

\begin{figure}[!t]
  \centering
  \includegraphics[width=\linewidth]{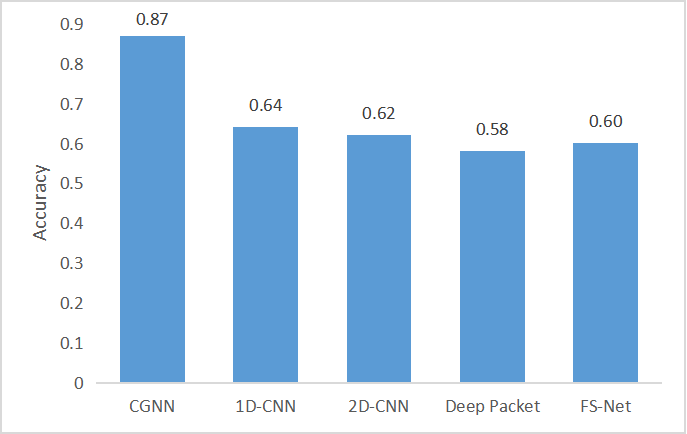}
  \caption{Prediction accuracy of different methods on the application traffic data sets.}
  \label{appacurracy}
\end{figure}

\textbf{Accuracy}: Figure \ref{appacurracy} shows the prediction accuracy for application categories. The accuracy of CGNN far exceeds the other methods by 23\% to 29\%. The significant improvement of the CGNN is mainly due to the fact that CGNN captures global structural features of the whole packet sequence.

\begin{figure}[!t]
  \centering
  \includegraphics[width=\linewidth]{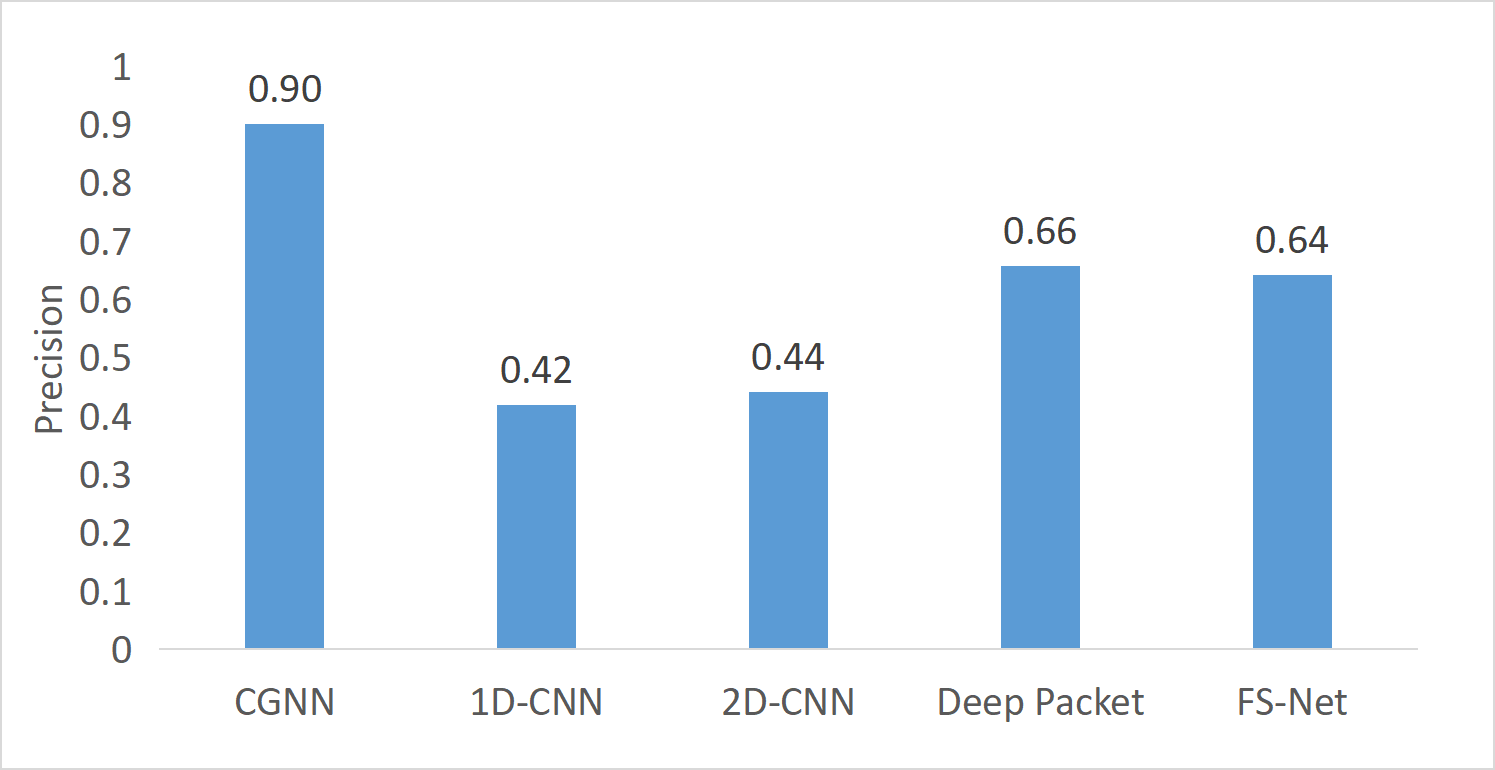}
  \caption{Average precision values of different methods on 41 kinds of application traffic data.}
  \label{precision_avg}
\end{figure}

\begin{figure}[!t]
  \centering
  \includegraphics[width=\linewidth]{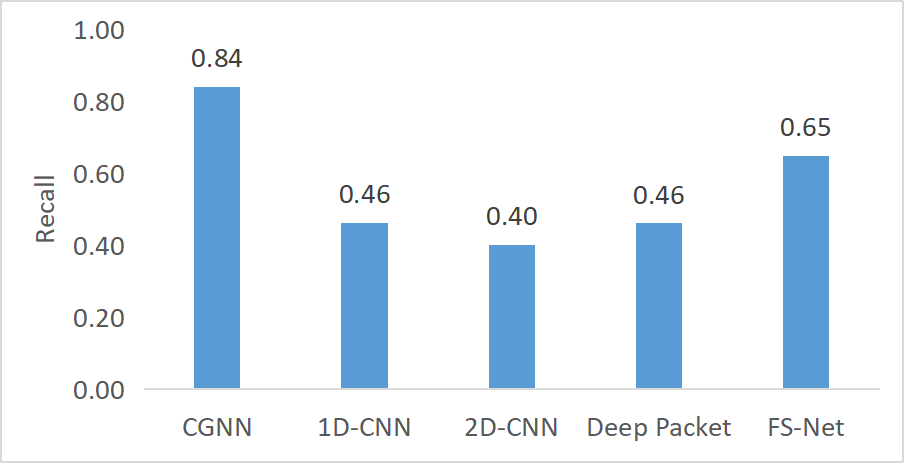}
  \caption{Average recall values of different methods on 41 kinds of application traffic data.}
  \label{recall_avg}
\end{figure}

\textbf{Recall and precision}: Having shown that CGNN is the most accurate, we next compare the recall and precision metrics. In order to visually show the difference between the classification performance of different methods, we calculated the average recall values and average precision values of each method on 41 applications. It can be seen from the Figure \ref{precision_avg} that compared with other methods, CGNN has increased the average precision value by 24\% to 48\%. And in the Figure \ref{recall_avg}, we can find that compared with other methods, CGNN improves the average recall value by 19\% to 44\%. Besides, we use the tables in the appendix to accurately compare each type of application data in order to highlight the gaps between different methods.

{\bfseries (ii) Experimental results of malicious traffic: }

\begin{figure}[!t]
  \centering
  \includegraphics[width=\linewidth]{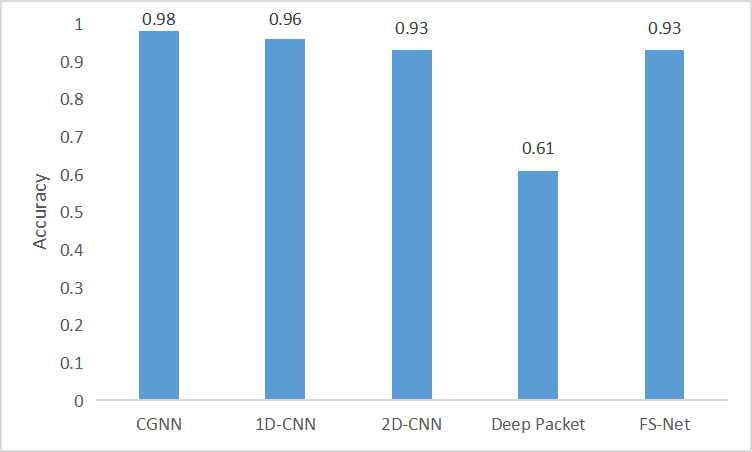}
  \caption{Prediction accuracy of different methods on the malicious traffic data sets.}
  \label{Maccuracy}
\end{figure}

\textbf{Accuracy}: Figure \ref{Maccuracy} shows that among these five models, the CGNN model has the highest accuracy. It can be seen that the characteristics of malicious traffic data are more distinguishable. 1D-CNN is the second best, while 2D-CNN and FS-Net are less accurate than CGNN and 1D-CNN. Deep Packet is the most inaccurate one, since it just outputs the classification result with few packets. In our experiments, CGNN improves the classification accuracy by 2\% to 37\%.

\begin{figure}[!t]
  \centering
  \includegraphics[width=\linewidth]{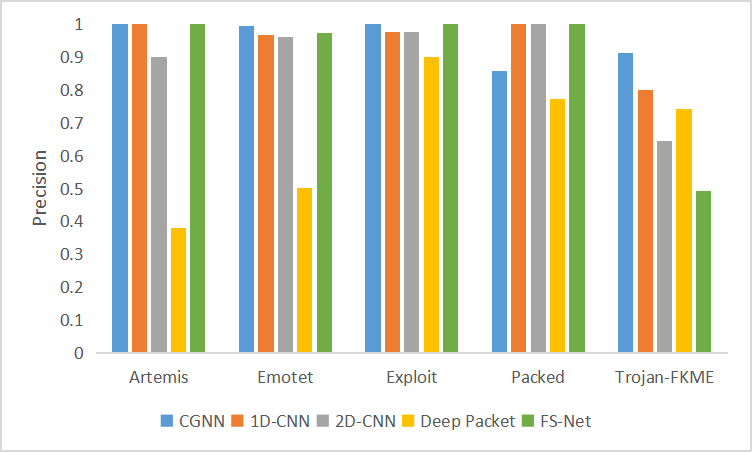}
  \caption{Precision values of different methods on 5 kinds of malicious traffic data.}
  \label{Mprecision}
\end{figure}

\begin{figure}[!t]
  \centering
  \includegraphics[width=\linewidth]{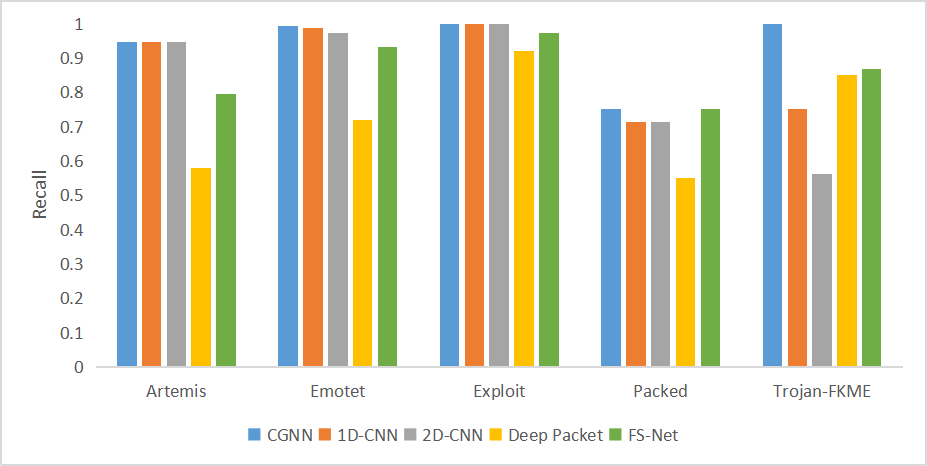}
  \caption{Recall values of different methods on 5 kinds of malicious traffic data.}
  \label{Mrecall}
\end{figure}

\textbf{Recall and precision}: Figure \ref{Mprecision} and Figure \ref{Mrecall} show the comparison of the recall and precision values of each type of malicious traffic after the 5 classification methods have classified 5 types of malicious traffic data. Among the 5 types of malicious traffic, we can see that the precision value of "Packed" in CGNN is slightly lower. The precision value of the other four malicious traffic is the highest in CGNN. At the same time, in the comparison results of recall, the results of CGNN are all the best. This shows that CGNN's classification of malicious traffic is relatively excellent, the advantages of the CGNN model are even greater. It is prominent and has relatively good classification performance for all types of malicious traffic. 

{\bfseries (iii) Experimental results of encrypted traffic: }

\begin{figure}[!t]
  \centering
  \includegraphics[width=\linewidth]{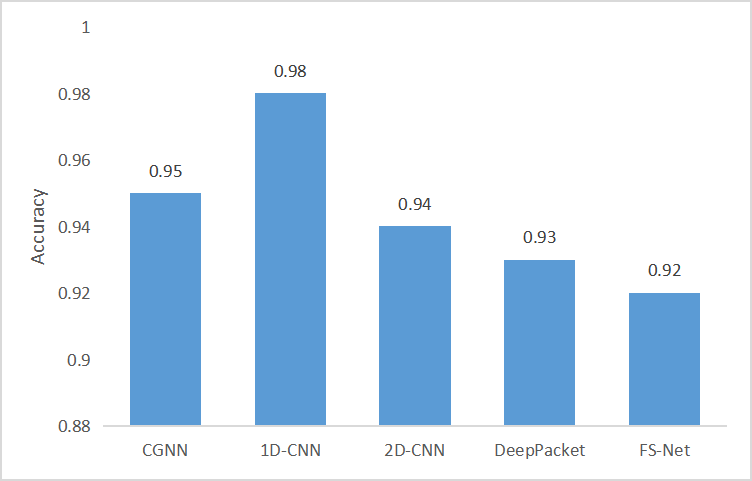}
  \caption{Prediction accuracy of different methods on  the  encrypted traffic data sets.}
  \label{entropyaccuracy}
\end{figure}

\textbf{Accuracy}: Figure \ref{entropyaccuracy} shows that the prediction accuracy of five methods  are at least 92\%. The prediction accuracy of 1D-CNN reaches 98\% and outperforms CGNN by 3\%. This is partly due to the fact that the average packet length of the encrypted packet is much smaller, while CGNN pads more zeros than 1D-CNN for each encrypted packet.

\begin{figure}[!t]
  \centering
  \includegraphics[width=\linewidth]{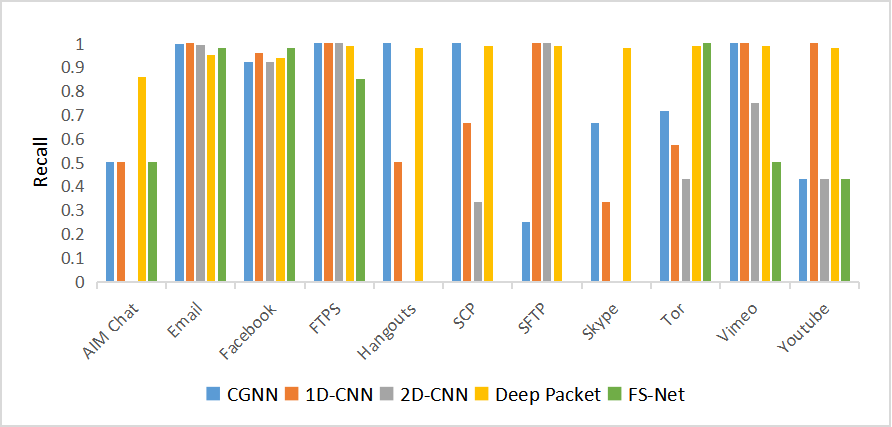}
  \caption{Recall values of different methods on 11 kinds of encrypted traffic data.}
  \label{entropyrecall}
\end{figure}

\begin{figure}[!t]
  \centering
  \includegraphics[width=\linewidth]{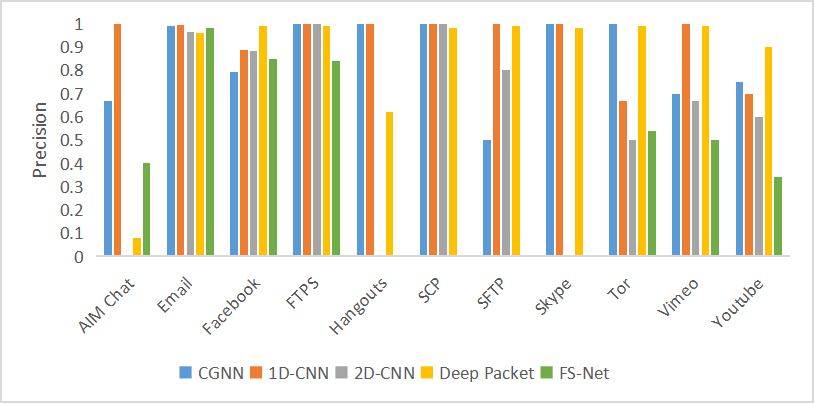}
  \caption{Precision values of different methods on 11 kinds of encrypted traffic data.}
  \label{entropyprecision}
\end{figure}

\textbf{Precision and recall}: Figures \ref{entropyrecall} and Figure \ref{entropyprecision} show the recall values and precision values obtained by the five methods on the encrypted traffic data set. In the classification of encrypted traffic, four methods (CGNN, Deep Packet, 1D-CNN, 2D-CNN) have better classification results. Although the accuracy values obtained by FS-Net looks pretty high, in Figure \ref{entropyrecall} and Figure \ref{entropyprecision}, the precision and recall values of FS-Net on Hangous, SCP, SFTP, and Skype are all 0. If the precision value of a type of traffic is 0, it means that no sample in the test set falls into that category, or the sample that falls into the traffic does not actually belong to the traffic. If the recall value is 0, it means that in fact, none of the samples belonging to this type of traffic have been correctly identified as this traffic. In short, if the accuracy or recall value of a traffic classification result is 0, it means that the classifier does not have the ability to correctly classify samples in the category. Therefore, we can see that FS-Net does not have the ability to classify these four types of encrypted traffic. Similarly, 2D-CNN will not have the ability to classify multiple encrypted traffic. In terms of the classification performance of encrypted traffic, the three models of CGNN, Deep Packet and 1D-CNN all have good classification capabilities for all types of encrypted traffic. Among them, 1D-CNN has the highest accuracy. In order to be more suitable for encrypting VPN traffic, the 1D-CNN model has adjusted the model structure and parameters. However, for non-VPN traffic, the accuracy of 1D-CNN will be reduced. The encrypted traffic used in this article is VPN traffic, so 1D-CNN has high accuracy.

{\bfseries (iv) Time complexity: } We next report the average training time and the average inference time. The table \ref{tab:time} shows the average time taken by different methods to classify the application dataset. In Table \ref{tab:time}, "avg. training time" represents the average training time used in a single epoch, "epoch" represents the number of rounds required to obtain the best result, "total.training time" represents the total training time, and "avg. inference time" represents the average inference time. The total training time of the CGNN method is shorter than Deep Packet and FS-Net, and longer than 1D-CNN and 2D-CNN, but the classification accuracy of 1D-CNN and 2D-CNN is very low.

\begin{table}[!t]
  \caption{Average time to perform training and that to infer the application label for an input.}
  \label{tab:time}
  \begin{tabular}{ccccc}
    \toprule
    Model & Per epoch &  Epochs & Training & Inference\\
    \midrule
    CGNN & 35s & 272&9520s & 1s\\
    Deep packet & 72s & 300 & 21600s & 110s\\
    1D-CNN & 0.05s & 40000& 2000s & 2s\\
    2D-CNN & 0.05s & 40000& 2000s &1s\\
    FS-Net & 396s & 150  &  59400s & 13s\\
    \bottomrule
  \end{tabular}
\end{table}

\subsection{Parameter Sensitivity}
\label{SenseSec}
After showing the comparison results, we can see that the classification performance of the CGNN method on various types of traffic data is better than the latest classification method in most cases. We next study the parameter sensitivity of the CGNN approach on the application traffic.  We choose the default parameter configuration in subsection \ref{paramSet}. The same conclusions hold for the other two data sets.

\begin{figure}[!t]
  \centering
  \includegraphics[width=\linewidth]{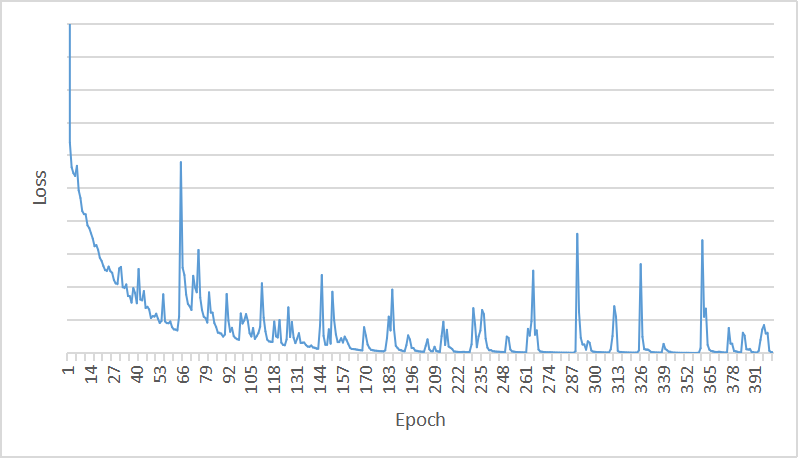}
  \caption{The convergence process of the loss value during the training process when classifying the application traffic data set.}
  \label{loss}
\end{figure}

{\bfseries (i) Convergence of the training: } During the training process, the convergence process of the loss value is shown in the Figure \ref{loss}. It can be seen in the figure that the CGNN model converges fast, and the diminishing returns occur after 100 epochs.

\begin{figure}[!t]
  \centering
  \includegraphics[width=0.7\linewidth]{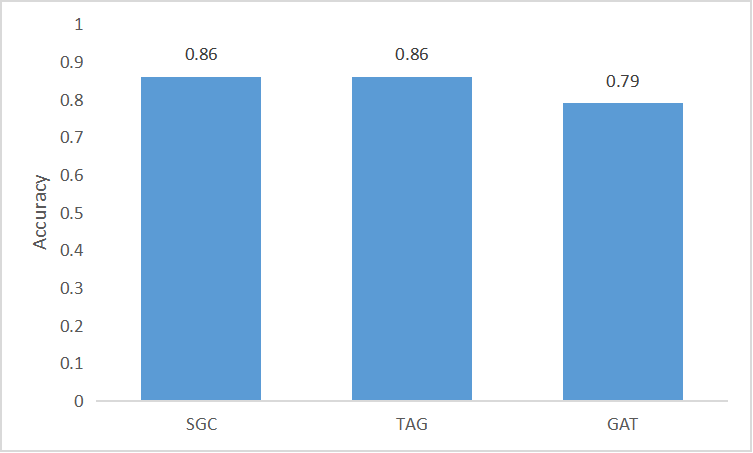}
  \caption{The impact of using different GNN models on classification accuracy}
  \label{modelaccu}
\end{figure}

{\bfseries (ii) Choice of neurons: } The graph neural network model supports common single-layer graph neural network models such as GCN \cite{kipf2016semi}, GAT \cite{velivckovic2017graph}, SGC \cite{wu2019simplifying}, TAG \cite{du2017topology}, etc. In these three alternative graph neural network frameworks, the experimental results of SGC and TAG are better,as shown in Figure \ref{modelaccu}, but relatively speaking, the running time of SGC is shorter and the efficiency is higher. 

{\bfseries (iii) Preprocessing or Not: } We next analyze whether transformation on the packet sequence provides any performance gains. The data preprocessing process is described in detail in subsection \ref{Preprocessing}. The preprocessing process converts a pcap file into a 1500-byte feature vector. Some other models may standardize the feature vector, that is, divide each byte by 255 to get a normalized vector with each value between 0.0 and 1.0. This preprocessing step seems to be relatively common when processing images, but it is not necessarily suitable for the payload of traffic data.

\begin{figure}[!t]
  \centering
  \includegraphics[width=0.6\linewidth]{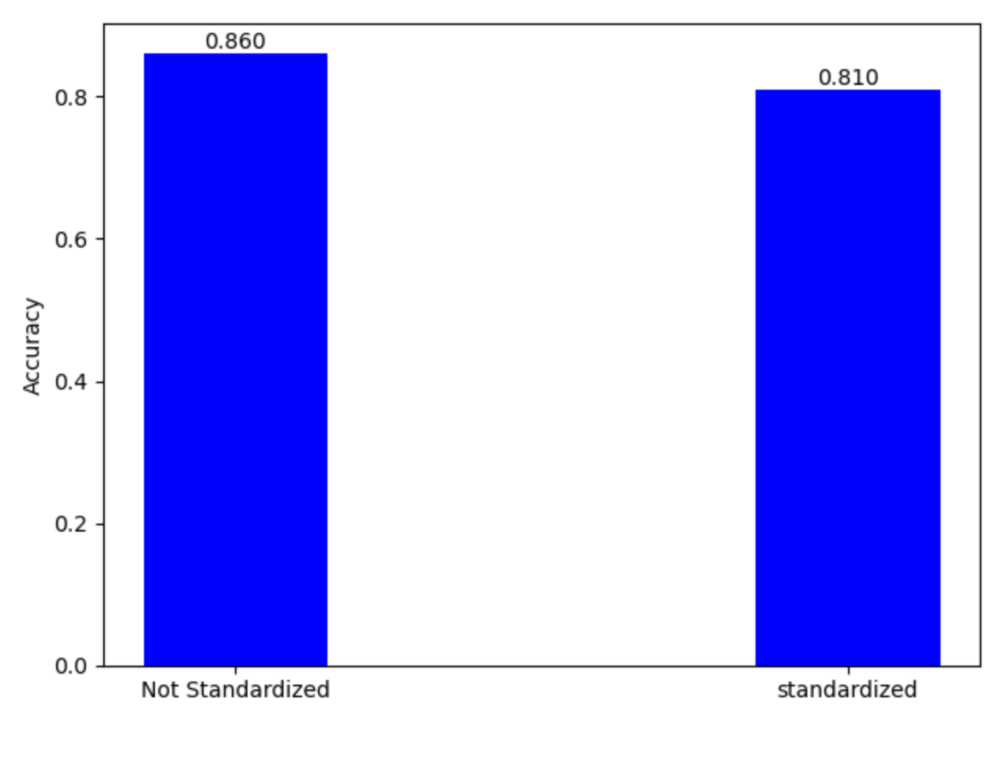}
  \caption{The accuracy after  preprocessing the pcap files and that without preprocessing them.}
  \label{standardized}
\end{figure}

Therefore, we do a comparative experiment on the preprocessed application dataset to compare the prediction accuracy obtained by the standardized packets and the non-standardized packets. In our experiments, we found that no standardization will lead to higher accuracy, as shown in the figure \ref{standardized}.
\co{We compare the prediction accuracy standardized packets with no-standardized packets.We found that the feature vector were standardized, but in our experiments we found that no standardization leads to a higher accuracy rate, as shown in Figure \ref{standardized}. }

{\bfseries (iv) The effect of the length of the feature vector: } Table \ref{pcaplengthtable} shows the comparison of accuracy rates obtained when 100, 500, 1000, and 1500 are selected as the byte length thresholds on the three data sets. We can see that choosing 1000 or 1500 as the threshold in the application data set is the best; choosing 500 as the threshold in the malicious data set is the highest, but in fact, the accuracy difference between the four thresholds is very small; It is best to use 100 as the threshold in the encrypted data set, but the difference between the three thresholds of 100, 1000 and 1500 is very small. On the whole, choosing 1500 as the threshold of the message byte length is the best choice.

\begin{table}[!t]
  \caption{Variations of the prediction accuracy as we change the length of the feature vector.}
  \label{pcaplengthtable}
  \begin{tabular}{ccccc}
    \toprule
    Dataset & 100 & 500 & 1000 & 1500\\
    \midrule
    application & 80.93\% & 85.62\% & 86.25\% & 86.37\% \\
    malware & 98.29\% & 98.60\% & 98.36\% & 98.36\% \\
   encrypted & 95.36\% & 94.07\% & 95.15\% & 95.23\% \\
    \bottomrule
  \end{tabular}
\end{table}

\begin{figure}[!t]
  \centering
  \includegraphics[width=\linewidth]{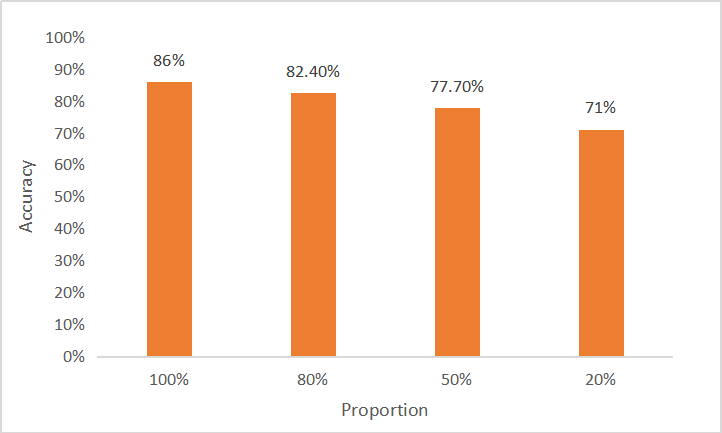}
  \caption{A comparison of the accuracy rates obtained by using different proportions of packets to generate a chained graph in a pcap file.}
  \label{Proportion}
\end{figure}

{\bfseries (v) Part or whole: } In addition, intuitively, it may be possible to use some packets in a pcap file to form a chained graph, thereby reducing the size of the chained graph. But according to experimental data,as shown in Figure \ref{Proportion}, the accuracy rate obtained by using the complete pcap chained diagram is the highest.

\begin{figure}[!t]
  \centering
  \includegraphics[width=\linewidth]{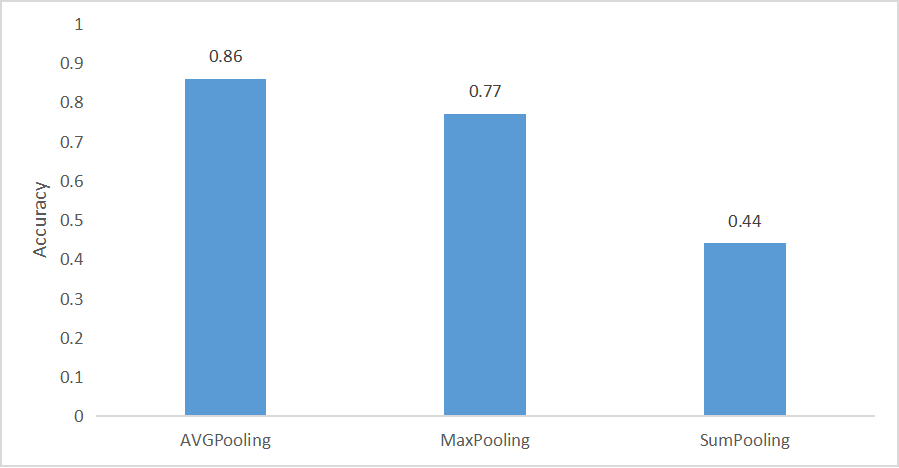}
  \caption{The impact of using different Pooling layers on classification accuracy.}
  \label{Pooling_acc}
\end{figure}

{\bfseries (vi) Pooling choice: }  In addition, we also tried adding different Pooling layers to the model. After testing, we found that AVGPooling has the best effect, as shown in Figure \ref{Pooling_acc}.

\begin{figure}[!t]
  \centering
  \includegraphics[width=\linewidth]{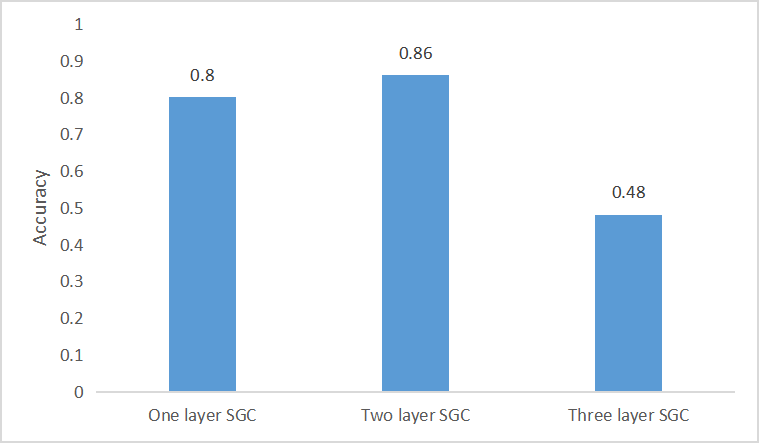}
  \caption{The impact of SGC with different layers on accuracy.}
  \label{diffSGC}
\end{figure}

{\bfseries (vii) Number of layers: }   In the CGNN model, we use a two-layer SGC because the graph neural network using the two-layer SGC model has good comprehensive accuracy and high training efficiency, as shown in Figure \ref{diffSGC}. Compared to using single-layer SGC, using double-layer SGC will improve classification performance. However, if a three-layer SGC is used, the classification effect will be greatly reduced due to the over-smoothing phenomenon \cite{2020DeeperGCN}.

\begin{figure}[!t]
  \centering
  \includegraphics[width=\linewidth]{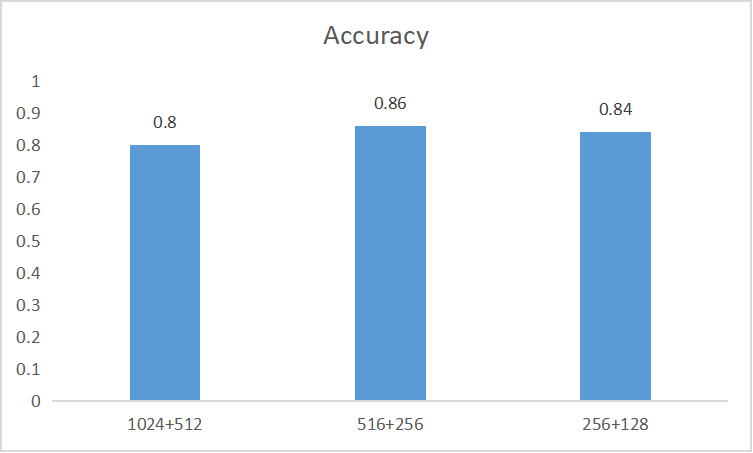}
  \caption{The output dimensions of the two-layer SGC have three different combinations (``516+256'' implies that each feature vector in the first layer is of length 516, and each feature vector in the second layer is of length 256.), and we compare their accuracy.}
  \label{SGC_parameters}
\end{figure}

{\bfseries (viii) The effect of SGC parameter combination on the classification result:}  We next compare the accuracy of the combination of feature numbers output by each layer. The first combination is that the output feature length of the first layer is 1024, and the output feature length of the second layer is 512; the second combination is that the output feature length of the first layer is 516, and the output feature length of the second layer is 256; the third combination The output feature length of the first layer is 256, and the output feature length of the second layer is 128. We found that in these three combinations, the first layer of neural network output feature length is 516, and the second layer of neural network output feature length is 256, the accuracy rate is the highest, as shown in Figure \ref{SGC_parameters}.

\co{
{\bfseries (x) Recommended parameters: }   Table \ref{cgnnparameters} describes the main parameters of each layer in the CGNN model based on the sensitivity analysis. 

\begin{table}[!t]
  \caption{THE MAIN PARAMETERS OF CGNN MODEL}
  \label{cgnnparameters}
  \begin{tabular}{ccccc}
    \toprule
    Layer & Operation & Input & k & Output\\
    \midrule
    1 & SGC+ReLU & 1500 & 1 & 516\\
    2 & SGC+ReLU & 516 & 1 & 256\\
    3 & AVGPooling & 256*n & -- & 256*1\\
    4 & full connect+softmax & 256 & -- & 41/5/11\\
    \bottomrule
  \end{tabular}
\end{table}
}

\section{Related Work}
\label{RelatedWorkSec}

 
Traffic classification has been extensively studied during the past decades. Here we report related studies that are most related to our work.  Lu et al. \cite{2016High} proposes two statistics-based solutions, the message size distribution classifier (MSDC) and the message size sequence classifier (MSSC) depending on classification accuracy and real timeliness. Sherry et al. proposed a deep data inspection system that can detect encrypted loads without decryption \cite{2015BlindBox}, but this system can only be effective for HTTPS traffic.

Traffic contents are increasingly used for classification due to their rich characteristics. Hajjar et al. \cite{2015Network}propose a novel blind, quintuple approach by exploring traffic attributes at the application level without inspecting the payloads. The identification model is based on the analysis of the first application-layer message based on their sizes, directions and positions in the flow. They think the first messages of a flow usually carry some application level signaling and data transfer units that can be discriminative through their patterns of size and direction. Wang et al. \cite{2015The} is the originator of the application of machine learning to traffic classification. The author believes that traffic characteristics can be regarded as pixels in images, so a better image classification program is used to classify traffic. Wang et al. \cite{2017End} proposed an end-to-end encrypted traffic classification method based on a one-dimensional convolutional neural network. This method is based on deep learning and uses 1D-CNN as a learning algorithm to automatically learn features directly from the original traffic. Lotfollahi et al. \cite{2020Deep} proposed a new method Deep Packet that uses deep learning to classify encrypted traffic. It proposed a deep learning-based method that integrates the feature extraction and classification stages into a system. After Deep Packet's initial data preprocessing stage, the data packet is input into the Deep Packet framework, which embeds a stacked autoencoder and convolutional neural network to classify traffic. Wang et al. \cite{2017Malware} and others input the traffic as a picture to the neural network for learning, that is, use 2D-CNN as a learning algorithm for traffic classification. In terms of graph representation of traffic flows, the traffic activity graph(TAG)  \cite{2009Unveiling} can capture application structures, where clients interact with servers and other clients. Its vertices represent IP addresses, edges represent traffic between IPs. The traffic causality graph(TCG) \cite{2014Network} has four types of relationships: (1) communication (CR); (2) propagation (PR); (3) dynamic-port host (DHR); and (4) static-port host (SHR). Its vertices and edges represent flows and flow relationships. The TCG-based models use SUBDUE’s minimum description length algorithm to convert the graphs to feature vectors based on discovered consistent substructures. The traffic dispersion graph(TDG)  \cite{2011Graption} can represent slices of TAG. It first groups flows using flow-level features, and uses TDGs to classify each group of flows. Different from prior studies that do not consider compositional and causal relationships in the packet sequence, this paper uses a chained graph to model these relationships, and perform classification on this chained graph model based on a novel graph neural network model to significantly improve the prediction performance.


\section{Conclusion}
\label{ConclusionSec}


We have proposed a novel chained graph model to preserve the structural and causal relationships in the network traffic, and  a graph neural network based traffic classifier, which classifies chained graphs to application categories. We perform extensive evaluation for CGNN with real-world application traffic, malicious traffic and encrypted traffic, and show that CGNN outperforms state-of-the-art neural network based traffic classifier. As our future work, we plan to integrate the CGNN model to real-world networking applications.

\newpage
\bibliographystyle{ACM-Reference-Format}
\bibliography{ref}

\newpage
\appendix

\section{Additional Experiment Results}

\subsection{Training and inference procedures}

 \begin{algorithm}[htbp]  
  \caption{ Training algorithm to optimize the parameters for the CGNN model based on samples of chained graphs. }  
  \label{alg:Framwork}  
  \begin{algorithmic}[1]  
    \Require  
      A train set of chained graphs $G_t$ for current batch. Each chained graph contains the feature vectors $X=[x_1,...,x_n]^T$ of all its nodes, graph structure matrix $A$ and the graph label $y_i$ of this graph;A valid set of chained graphs $G_v$.
      The list $Y=[y_1,y_2,...,y_m]$ of labels of chained graphs for current batch $G_t$ and the list $Y_v$ of labels of chained graphs for valid set$G_v$. Let $Y$ represent the true category of all chained graph samples, and $\hat{Y}$ represents the predicted category.
    \Ensure  
      The optimal classification model $M$ obtained on the input training set.
    \State Send $X$ and $A$ to the first layer of SGC, the input feature dimension of SGC is $p$, and the output dimension is $d_1$. So we can get the output feature $\overline{X}^{(1)}$ of shape ($N$, $d_1$) where $d_1$ is size of output feature and $N$ is the number of nodes.
    
    $\overline{X}^{(1)}=SGC_1(p,d_1,X,A)$.
    \label{code:fram:extract}  
    \State Send $\overline{X}^{(1)}$ and $A$ to the second layer of SGC, the input feature dimension of SGC is $d_1$, and the output dimension is $d_2$. So we can get the output feature matrix $\overline{X}^{(2)}$ of shape ($N$, $d_2$) where $d_2$ is size of output feature and $N$ is the number of nodes.  
    
    $\overline{X}^{(2)}=SGC_2(d_1, d_2,\overline{X}^{(1)},A)$.
    \label{code:fram:trainbase}  
    \State Send $\overline{X}^{(2)}$ to the layer of AVGPooling. The output feature $y_i^{(1)}$ is the pooling result of $g_i$ which is a chained graph in the current graph set $G$.
    
    $g_i=AVGPooling(\overline{X}^{(2)})$ where $G=[g_1,g_2,...,g_n]$
    \label{code:fram:trainbase}
    \State Send $y_i^{(1)}$ to the output layer. The input of the layer is a $d_2$-byte graph feature vector. $y_i^{(1)}$ is sent to the linear layer to get the feature vector $y_i^{(2)}$ of shape (1,$k$) where $k$ is the number of kinds. Send $y_i^{(2)}$ to the softmax function, and then get the classification result  for current batch $\hat{Y}=[\hat{y}_1,...,\hat{y}_m]$ where $\hat{y}=softmax(y_i^{(2)})$ is the classification result of $g_i$. 
    
    $y_i^{(2)}=Linear(d_2,k,y_i^{(1)},A)$ and then $\hat{y_i}=softmax(y_i^{(2)})$. Therefore we can get $\hat{Y}=[\hat{y_1},\hat{y_2},...,\hat{y_n}]$
    \label{code:fram:add}  
    \State Use the logistic regression function to calculate the loss value between the predicted labels $\hat{Y}$ and the actual labels $Y$.
    
    $loss=CrossEntropyLoss(\hat{Y},Y)$
    \label{code:fram:classify}  
    \State Initialize the gradient weight to 0; propagate the calculated loss backwards through the optimizer to obtain a new gradient and update all parameters.

    \label{code:fram:select} 
    \State After each epoch of training on the training set, the classification model parameters are fixed and the valid chained graph set $G_v$ are send to it to get the predicted labels $\hat{Y}_v$.
    \label{code:fram:select}   
    \State Calculate the loss between the real labels $Y_v$ and the predicted labels $\hat{Y}_v$. If the loss is small enough and does not decrease, stop training the model, otherwise, it will return to step 1 to start a new round of training.
    \label{code:fram:select} \\  
    \Return Classification model $M$ and its parameters.
  \end{algorithmic}  
\end{algorithm} 

Algorithms  \ref{alg:Framwork}   and \ref{alg:InferenceFramwork} show the details of the training and inference processes.

 
  \begin{algorithm}[htbp]  
  \caption{The inference algorithm to obtain the classification result for a list of input chained graphs.}  
  \label{alg:InferenceFramwork}  
  \begin{algorithmic}[1]  
    \Require  
      A test set of chained graphs $G_{test}$, each graph has its feature matrix $X$ and its structure matrix $A$.
      The list $Y_{test}=[y_1,y_2,...,y_m]$ of labels of chained graphs for test set.
    \Ensure  
      The predicted graph label list $\hat{Y}_{test}$.
    \State Send the test set $G_{test} $ to the trained model. Like the first step in the training algorithm, then we can obtain $\overline{X}^{(1)}$ in the first layer of SGC.
    \label{code:fram:input}  
    \State Send $\overline{X}^{(1)}$ and $A$ to the second layer of SGC, the input feature dimension of SGC is $d_1$, and the output dimension is $d_2$. So we can get the output feature $\overline{X}^{(2)}$ of shape ($N$,$d_2$) where $d_2$ is size of output feature and $N$ is the number of nodes.  
    \label{code:fram:trainbase}  
    \State Send $\overline{X}^{(2)}$ to the layer of AVGPooling. The output feature $y_i^{(1)}$ is the pooling result of $g_i$ which is a chained graph in the test set $G_{test}$.  
    \label{code:fram:trainbase}
    \State Send $y_i^{(1)}$ to the output layer. The input of the layer is a $d_2$-byte graph feature vector. $y_i^{(1)}$ is sent to the linear layer to get the feature vector $y_i^{(2)}$ of shape (1,$k$) where $k$ is the number of kinds. Send $y_i^{(2)}$ to the softmax function, and then get the classification result  for test set $\hat{Y}_{test}=[\hat{y}_1,...,\hat{y}_m]$ where $\hat{y}=softmax(y_i^{(2)})$ is the classification result of $g_i$. 
    \label{code:fram:add}\\
    \Return Output the predicted graph label list $\hat{Y}_{test}$.
  \end{algorithmic}  
\end{algorithm}

\subsection{Data Set Analysis}

\begin{figure}[!t]
  \centering
  \includegraphics[width=\linewidth]{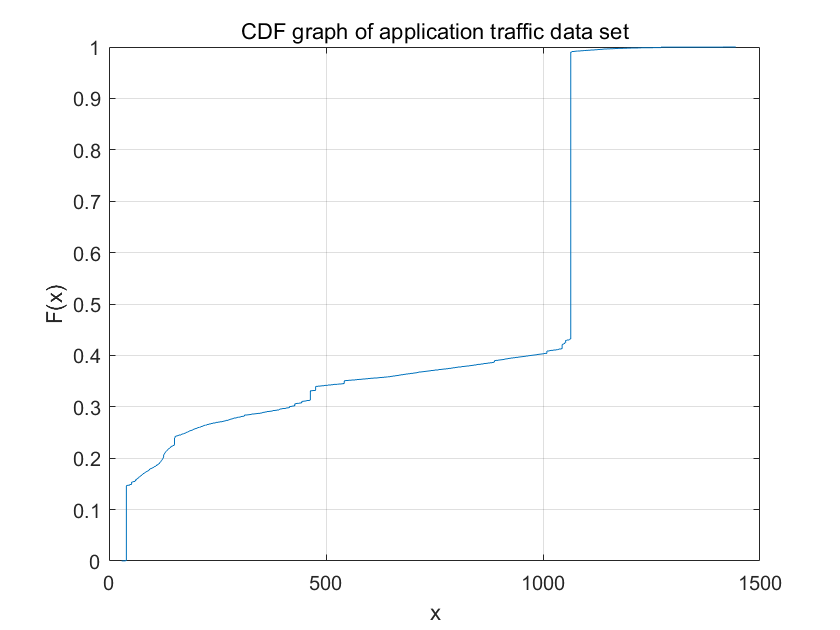}
  \caption{CDF graph of application traffic data set.}
  \label{ACDF}
\end{figure}

\begin{figure}[!t]
  \centering
  \includegraphics[width=\linewidth]{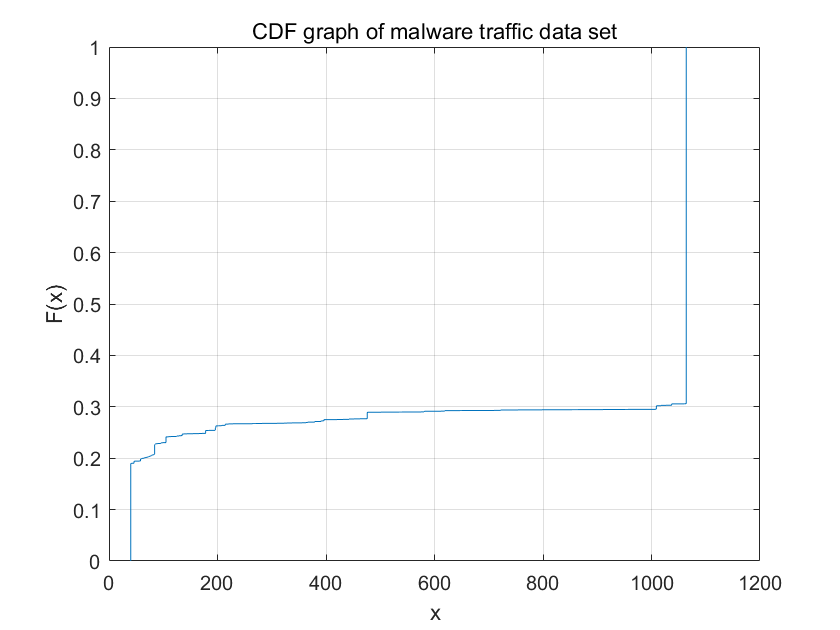}
  \caption{CDF graph of malware traffic data set.}
  \label{MCDF}
\end{figure}

\begin{figure}[!t]
  \centering
  \includegraphics[width=\linewidth]{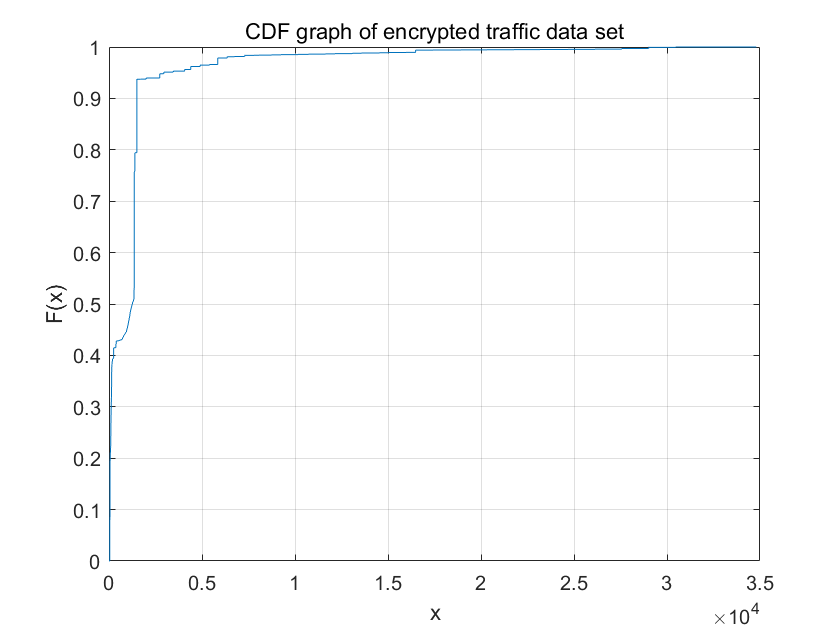}
  \caption{CDF graph of encrypted traffic data set.}
  \label{ECDF}
\end{figure}

Figures \ref{ACDF} to Figure \ref{ECDF} respectively show the cumulative distribution function(CDF) graphs of the application data set, the malicious traffic data set and the encrypted data set. The horizontal axis represents the length of the packets, and the vertical axis represents the cumulative distribution. As we can see in the images, the packet lengths in the application data set are all below 1500, and most of the data packets are below 1200. In the malicious traffic data set, the packet lengths are all below 1200, and most of the packet lengths are below 1100. In the encrypted traffic data set, the packet lengths are all below 3500, most of which are below 1500. In such a distribution situation, we choose 1500 as the threshold and intercept the first 1500 bytes of these packets. If they are less than 1500 bytes, we pad zeros to keep them to be of length 1500 bytes. 

\subsection{Heat Map of Traffic Classification Results for CGNN}

\begin{figure}[!t]
  \centering
  \includegraphics[width=\linewidth]{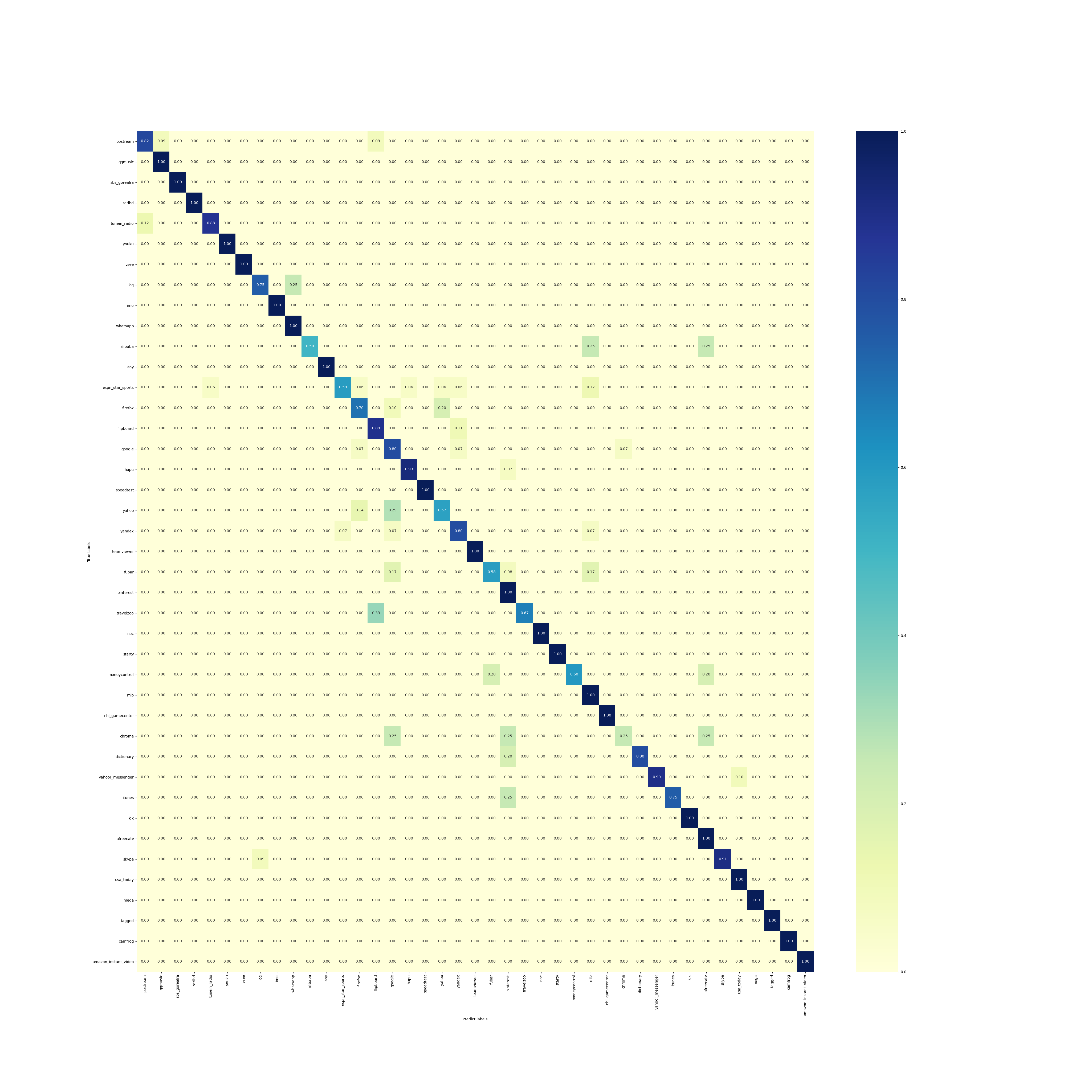}
  \caption{A heat map of the classification results of CGNN classifying application traffic.}
  \label{appmatrix1}
\end{figure}

We study the classification sensitivity of CGNN on different data sets. From Figures \ref{appmatrix1}, we see that the diagonal entries are close to the optimal in most cases. The inaccurate classification results are typically asymmetric. The heat map shows that the classification result of application classifier for most applications is relatively good, with strong discrimination.

\begin{figure}[!t]
  \centering
  \includegraphics[width=\linewidth]{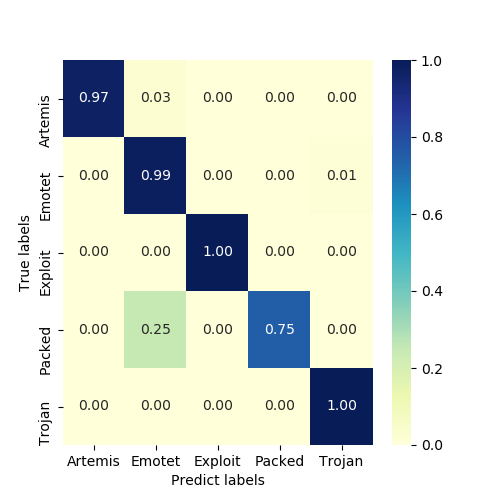}
  \caption{Heatmap of CGNN on the malicious traffic.}
  \label{mheatmap2}
\end{figure}

\begin{figure}[!t]
  \centering
  \includegraphics[width=\linewidth]{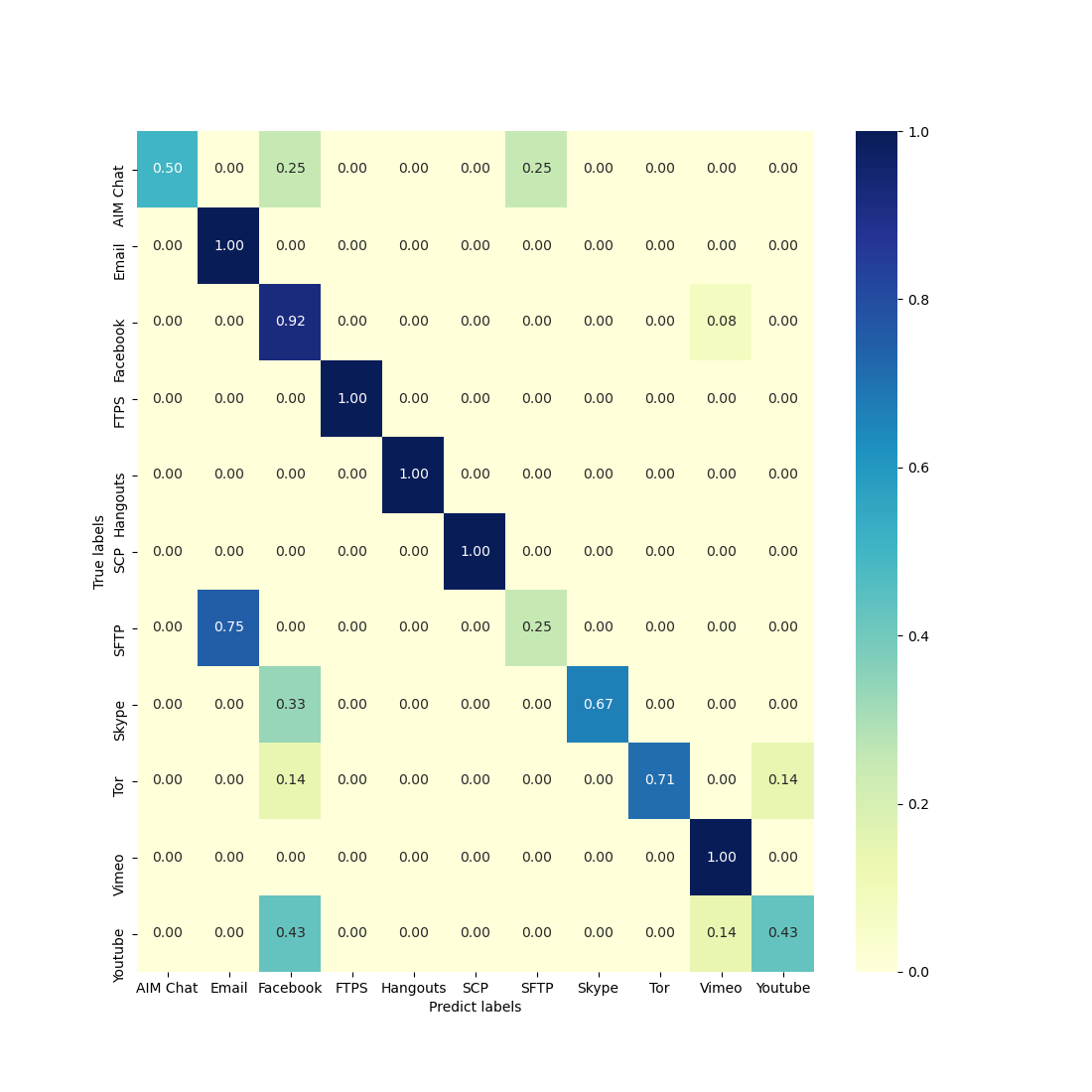}
  \caption{Heatmap of CGNN on the encrypted traffic.}
  \label{entropy_heatmap3}
\end{figure}

Figure \ref{mheatmap2} shows the heat map of the CGNN model to classify malicious traffic.It can also be seen from the heat map that GCNN has such a high classification accuracy for 5 types of malicious traffic. Figure \ref{entropy_heatmap3} shows the heat map of the CGNN model to classify encrypted traffic.

\end{document}